\documentclass[conference]{IEEEtran}
\IEEEoverridecommandlockouts
\usepackage{cite}
\usepackage{amsmath,amssymb,amsfonts}
\usepackage{amsthm}
\usepackage{graphicx}
\usepackage{textcomp}
\usepackage{xcolor}
\usepackage{subfigure}

\usepackage{booktabs}
\usepackage{algorithm}
\usepackage{amsmath}
\usepackage[noend]{algpseudocode}
\usepackage{multirow}
\usepackage{bm}
\usepackage{amsbsy}
\usepackage{scalerel}
\usepackage{tabu}

\usepackage{hyperref}
\usepackage{cleveref}

\usepackage{multirow}
\usepackage{tabulary}
\newcolumntype{C}[1]{>{\centering\arraybackslash}p{#1}}

\def\BibTeX{{\rm B\kern-.05em{\sc i\kern-.025em b}\kern-.08em
    T\kern-.1667em\lower.7ex\hbox{E}\kern-.125emX}}
\begin{document}

\title{Sequence-Aware Factorization Machines \\for Temporal Predictive Analytics}

%\{xueli,zxf\}@itee.uq.edu.au}

\author
{
 Tong Chen{\small$^\dag$}\hspace*{10pt}Hongzhi Yin{\small$^\dag$}$^*$\thanks{$^*$Corresponding author; contributing equally with the first author.\newline2020 IEEE 36th International Conference on Data Engineering (ICDE)}\hspace*{10pt}Quoc Viet Hung Nguyen{\small$^\ddag$}\hspace*{10pt}Wen-Chih Peng{\small$^\S$}\hspace*{10pt}Xue Li{\small$^\dag$}\hspace*{10pt}Xiaofang Zhou{\small$^\dag$}\\
 \fontsize{10}{10}\selectfont\itshape $~^\dag$School of Information Technology and Electrical Engineering, The University of Queensland,\\
 {\fontsize{9}{9}\selectfont\ttfamily\upshape \{tong.chen,h.yin1\}@uq.edu.au} \hspace{10pt}{\fontsize{9}{9}\selectfont\ttfamily\upshape \{xueli,zxf\}@itee.uq.edu.au}\\
   \fontsize{10}{10}\selectfont\itshape $~^\ddag$School of Information and Communication Technology, Griffith University, {\fontsize{9}{9}\selectfont\ttfamily\upshape quocviethung.nguyen@griffith.edu.au}\\
  \fontsize{10}{10}\selectfont\itshape $~^\S$Department of Computer Science, National Chiao Tung University, {\fontsize{9}{9}\selectfont\ttfamily\upshape wcpeng@g2.nctu.edu.tw}\\
  %[-3.0ex]
}
\maketitle

\begin{abstract}
In various web applications like targeted advertising and recommender systems, the available categorical features (e.g., product type) are often of great importance but sparse. As a widely adopted solution, models based on Factorization Machines (FMs) are capable of modelling high-order interactions among features for effective sparse predictive analytics. As the volume of web-scale data grows exponentially over time, sparse predictive analytics inevitably involves dynamic and sequential features. However, existing FM-based models assume no temporal orders in the data, and are unable to capture the sequential dependencies or patterns within the dynamic features, impeding the performance and adaptivity of these methods. Hence, in this paper, we propose a novel Sequence-Aware Factorization Machine (SeqFM) for temporal predictive analytics, which models feature interactions by fully investigating the effect of sequential dependencies. As static features (e.g., user gender) and dynamic features (e.g., user interacted items) express different semantics, we innovatively devise a multi-view self-attention scheme that separately models the effect of static features, dynamic features and the mutual interactions between static and dynamic features in three different views. In SeqFM, we further map the learned representations of feature interactions to the desired output with a shared residual network. To showcase the versatility and generalizability of SeqFM, we test SeqFM in three popular application scenarios for FM-based models, namely ranking, classification and regression tasks. Extensive experimental results on six large-scale datasets demonstrate the superior effectiveness and efficiency of SeqFM. 

\end{abstract}

\section{Introduction}\label{sec:intro}
As an important supervised learning scheme, predictive analytics play a pivotal role in various applications, ranging from recommender systems \cite{wu2017recurrent,yang2017bridging} to financial analysis \cite{chen2018tada} and online advertising \cite{juan2016field,zhou2018deep}. In practice, the goal of predictive analytics is to learn a mapping function from the observed variables (i.e., features) to the desired output. 

When dealing with categorical features in predictive analytics, a common approach is to convert such features into one-hot encodings \cite{he2017neuralcol,shan2016deep,rendle2011fast} so that standard regressors like logistic regression \cite{hosmer2013applied} and support vector machines \cite{chang2011libsvm} can be directly applied. Due to the large number of possible category variables, the converted one-hot features are usually of high dimensionality but sparse \cite{he2017neural}, and simply using raw features rarely provides optimal results. On this occasion, the interactions among different features act as the winning formula for a wide range of data mining tasks \cite{lian2017practical,yin2017sptf,shan2016deep}. The interactions among multiple raw features are usually termed as \textbf{\textit{cross features}} \cite{shan2016deep} (a.k.a. multi-way features and combinatorial features). For example, individual variables $occupation = \{lecturer, engineer\}$ and $level = \{junior, senior\}$ can offer richer contextual information for user profiling with cross features, such as $(junior, engineer)$ and $(senior, lecturer)$. To avoid the high cost of task-specific manual feature engineering, \textbf{\textit{factorization machines}} (FMs) \cite{rendle2010factorization} are proposed to embed raw features into a latent space, and model the interactions among features via the inner product of their embedding vectors. 

To better capture the effect of feature interactions, variants of the plain FM are proposed, like field-aware FM for online advertising \cite{juan2016field} and CoFM \cite{hong2013co} for user behavior modelling. However, these variants are still constrained by their limited linear expressiveness \cite{he2017neural} when modelling the subtle and complex feature interactions. Recently, motivated by the capability of learning discriminative representations from raw inputs, deep neural networks (DNNs) \cite{lecun2015deep} have been adopted to extend the plain FM. For instance, He \textit{et al.} \cite{he2017neural} bridges the cross feature scheme of FM with the non-linear form of DNN, and proposes a neural factorization machine (NFM). Instead of the straightforward inner product in FM, NFM takes the sum of all features' linear pairwise combinations into a feed-forward neural network, and generates a latent representation of high-order feature interactions.  With the idea of learning high-order feature interactions with DNNs, various DNN-based FMs are devised for predictive analytics \cite{xiao2017attentional,cheng2016wide,lian2018xdeepfm,shan2016deep,guo2017deepfm,qu2016product}. 

In short, there are two major trends of improvements over the plain FM. One is to make the model ``deep" with multi-layer network structures in order to exhaustively extract useful information from feature interactions, e.g., the residual network in DeepCross \cite{shan2016deep}, the pairwise product layer in PNN \cite{qu2016product}, and the compressed interaction network in xDeepFM \cite{lian2017practical}. The other is to make the model ``wide" by considering multiple feature interactions in varied domains (usually coupled with ``deep" structures), e.g., separately modelling user logs and texts with CoFM \cite{hong2013co}, or fusing shallow low-order output with dense high-order output via Wide\&Deep \cite{cheng2016wide}, DeepFM \cite{guo2017deepfm} and xDeepFM \cite{lian2018xdeepfm}. Note that in the remainder of this paper, to avoid ambiguity, we use the term \textbf{\textit{FM-based models}} to imply both the plain FM and all its variants.

However, these popular FM-based models mostly perform predictive analytics with the assumption that there is no temporal order in the data. As a result, regardless of the temporal information available in various prediction tasks, the data will be partitioned for training/evaluation randomly rather than chronologically, such as \cite{rendle2011fast,he2017neural,xiao2017attentional,rendle2012factorization}. Considering a real-world recommendation scenario, the time-dependent order of products purchased by each user should be considered, and the recommender system can only utilize users' past purchase records to estimate their future preferences \cite{campos2014time}. To this end, we focus on the problem of \textbf{\textit{temporal predictive analytics}} which considers such temporal causality, and is more practical and realistic in various application scenarios.

\begin{figure}[!t]
\center
\includegraphics[width = 3.5in]{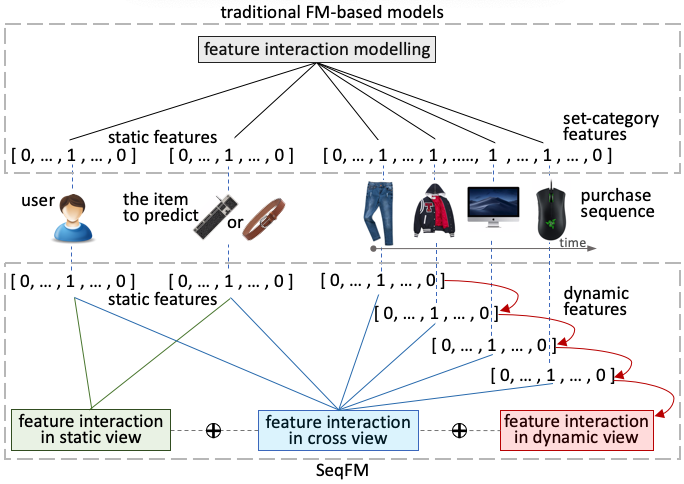}
\vspace{-0.8cm}
\caption{The differences in feature interaction modelling between traditional FM-based models (upper part) and our proposed SeqFM (lower part). Note that the embedding process of sparse features is omitted to be succint.}
\label{Figure:Struct}
\vspace{-0.5cm}
\end{figure}

Despite the efforts on enhancing the plain FM, all the aforementioned FM-based models still lack the consideration of the sequential dependencies within high-order feature interactions, which is proven to be critical for many temporal prediction tasks \cite{lai2017modeling,chen2018tada,kang2018self,wang2016spore,guo2019streaming}.  
With the rapidly increasing volume of web-scale data, temporal predictive analytics inevitably involves features that are dynamically changing over time, e.g., users' shopping transactions on e-commerce platforms. We term such features the \textbf{\textit{dynamic features}}. In contrast, we refer to features that stay fixed (e.g., user ID) as \textbf{\textit{static features}}. Let us consider a generic item recommendation task, where the goal is to predict whether a user will buy a specific item or not, as shown in Figure \ref{Figure:Struct}. Apart from the one-hot encoding of both the user and candidate item, the common way for current FM-based models to account for this user's shopping record is to derive set-category features \cite{rendle2011fast,zhou2018deep,rendle2012factorization} that mark all her/his previously bought items (see Figure~\ref{Figure:Struct}). As is inferred from the user's transaction (jeans $\!\rightarrow\!$ jacket $\!\rightarrow\!$ computer $\!\rightarrow\!$ mouse), the current intent of this user is to purchase accessories for her/his new computer like keyboards, rather than other clothes. However, since traditional FM-based models view all the purchased items from a constant point of time, all these dynamic features are evenly treated when modelling feature interactions. Consequently, traditional FM-based models can hardly distinguish the likelihood of purchasing a keyboard with purchasing a belt, because there are similar items in the set-category features for both keyboards and belts, and the sequential characteristics of dynamic features cannot be properly captured. Though the recently proposed translation-based FM \cite{pasricha2018translation} performs recommendation by taking the sequential property of features into account, it models the influence of only the last item (i.e., the mouse), thus easily making the recommended keyboard a mismatch for the purchased computer. Moreover, for FM-based models, the deficiency of handling sequential dependencies will create a severe performance bottleneck when the diversity and amount of dynamic features grow over time.

In light of this, we aim to develop a general yet effective FM-based model to thoroughly mine the sequential information from the dynamic features for accurate temporal predictive analytics. Hence, in this paper, we propose a \textbf{\textit{Sequence-Aware Factorization Machine}} (SeqFM), which is the first FM-based model to systematically combine sequential dependencies with feature interactions while inheriting the non-linear expressiveness from DNNs and retaining the compactness w.r.t. the plain FM. As demonstrated in Figure \ref{Figure:Struct}, SeqFM is built upon a multi-view learning scheme. Due to different semantic meanings carried by static and dynamic features, we model different types of feature interactions in three different contexts (i.e., views): static view for static features, dynamic view for dynamic features, and cross view for both. To bypass the high demand on space and time of sequential neural models using convolutional or recurrent computations, in each specific view, we leverage the self-attention mechanism \cite{vaswani2017attention}, which is highly efficient and capable of uncovering sequential and semantic patterns between features. For the dynamic view and cross view, we further propose two masked self-attention units to respectively preserve the directional property of feature sequence and block irrelevant feature interactions. After encoding the high-order interactions between features via the multi-view self-attention, a shared residual feed-forward network is deployed to extract latent information from feature interactions. 

Intuitively, compared with ``deep" or ``wide" FM variants, we aim to make our model ``sequence-aware", thus making full use of the contexts within dynamic features. As a flexible and versatile model, we introduce three application scenarios for SeqFM, namely ranking, classification, and regression, where corresponding experiments reveal significant improvements over existing FM-based models. Furthermore, the simple structure of SeqFM also ensures linear computational complexity and light-weight parameter size.

In summary, our work contributes in the following aspects:
\begin{itemize}
	\item We point out that mining features' sequential dependencies can greatly benefit the modelling of feature interactions in real-world FM-based models. We introduce, to the best of our knowledge, the first study to endow FM-based models with full sequence-awareness for temporal predictive analytics.    
	\item We propose SeqFM, a novel sequence-aware factorization machine. SeqFM utilizes an innovative multi-view self-attention scheme to model the high-order feature interactions in a sequence-aware manner.
	\item We conduct extensive experiments on a wide range of benchmark datasets to showcase the superiority of SeqFM in different temporal predictive analytic tasks, validate the importance of sequence-awareness in SeqFM, and reveal promising practicality and scalability of SeqFM. 
\end{itemize}

\section{Preliminaries}\label{sec:pre}
\textbf{Notations.} Throughout this paper, all vectors and matrices are respectively denoted by bold lower case and bold upper case letters, e.g., $\textbf{g}$ and $\textbf{G}$. All vectors are \textbf{row vectors} unless specified, e.g., $\textbf{x} \in \mathbb{R}^{1\times n}$. To maintain simplicity, we use the superscripts $\circ$, $\triangleright$ and $\star$ to distinguish parameters in the static view, dynamic view and cross view, respectively.

\textbf{Factorization Machines (FMs).}
FMs are originally proposed for collaborative recommendation. Specifically, for a given instance $[user \, ID\!=\!2, gender\!=\!male, cities \,\, visited\!= Sydney \& Shanghai]$, its input is a high-dimensional sparse feature $\textbf{x} \in \{0,1\}^{1\times m}$ constructed by the concatenation of multiple one-hot encodings \cite{lian2018xdeepfm,shan2016deep,qu2016product}:
\vspace{-0.1cm}
\begin{equation}\label{eq:input}
	\textbf{x} = \underbrace{[0,1,0,...,0]}_\textnormal{\small user ID} \underbrace{[1,0]}_\textnormal{\small gender} \underbrace{[0,1,0,1,0,...,0]}_\textnormal{\small cities visited},
\vspace{-0.1cm}
\end{equation}
where any real-valued feature (e.g., age) can also be directly included in $\textbf{x}$ \cite{he2017neural,rendle2011fast}, but we will focus on the sparse categorical feature in our paper. Then, FMs are linear predictors that estimate the desired output by modelling all interactions between each pair of features within $\textbf{x}$ \cite{rendle2010factorization}:
\vspace{-0.1cm}
\begin{equation}\label{eq:FM}
	\widehat{y} = w_0 + \sum^m_{i=1}{w_ix_i} + \sum^m_{i=1}\sum^m_{j=i+1}{\langle \textbf{v}_i , \textbf{v}_j \rangle \cdot x_ix_j},
\vspace{-0.1cm}
\end{equation}
where $m$ is the total amount of features, $w_0$ is the global bias, $w_i$ is the weight assigned to the $i$-th feature, and $\langle .,.\rangle$ denotes the dot product of two vectors. $\textbf{v}_i$, $\textbf{v}_j \!\in\! \mathbb{R}^{1\times d}$ are corresponding embedding vectors for feature dimension $i$ and $j$, while $d$ is the embedding dimension. Thus, the first two terms in Eq.(\ref{eq:FM}) can be viewed as a linear weighting scheme, while the third term models the effect of pairwise feature interactions \cite{xiao2017attentional}.

\section{Sequence-Aware Factorization Machines}\label{sec:model}
In this section, we first overview our proposed Sequence-Aware Factorization Machines (SeqFM), and then detail each key component in the model.

Given a sparse feature vector $\textbf{x} \in \{0,1\}^{1\times m}$, the output $\widehat{y}$ of SeqFM is computed via:
\vspace{-0.1cm}
\begin{equation}\label{eq:SeqFM_OG}
	\widehat{y} = w_0 + \sum^m_{i=1}{w_ix_i} + f(\textbf{x}),
\vspace{-0.1cm}
\end{equation}
where the first two terms denote the linear components similar to the ones in Eq.(\ref{eq:FM}), and the global bias and weights of different features are modelled respectively. $f(\textbf{x})$ denotes our proposed factorization component. Based on the construction rule of $\textbf{x}$, it can be viewed as the additive form of the one-hot encodings for all non-zero features. Thus, $\textbf{x} = \sum _{i=1}^{n}{\textbf{g}_i}$, where $\textbf{g}_i = [0,...,0,1,0,...,0]$ is an $m$-dimensional one-hot vector corresponds to one individual non-zero feature, and $n$ denotes the total number of non-zero features.

To conduct temporal predictive analytics with sequence-awareness, we split the original sparse feature vector $\textbf{x}$ into two views, namely the \textit{\textbf{static view}} and \textit{\textbf{dynamic view}}. In the running example of Eq.(\ref{eq:input}), user ID and gender are modelled in the static view while visited cities are modelled in the dynamic view. Then, we can obtain the static feature $\textbf{x}^{\circ} \in \{0,1\}^{1\times m^{\circ}}$ and dynamic feature $\textbf{x}^{\triangleright} \in \{0,1\}^{1\times m^{\triangleright}}$ where $m^{\circ} + m^{\triangleright} = m$. 
Correspondingly, the additive form of input features naturally splits into $\textbf{x}^{\circ} \!\!=\!\! \sum _{i=1}^{n^{\circ}}{\textbf{g}_i^{\circ}}$ and $\textbf{x}^{\triangleright} \!\!=\!\! \sum _{i=1}^{n^{\triangleright}}{\textbf{g}_i^{\triangleright}}$, where $n^{\circ}$ and $n^{\triangleright}$ are the respective numbers of non-zero features in two views, and $n^{\circ} + n^{\triangleright} = n$. Here, we use feature matrices $\textbf{G}^{\circ} \in \{0,1\}^{n^{\circ} \times m^{\circ}}$ and $\textbf{G}^{\triangleright} \in \{0,1\}^{n^{\triangleright} \times m^{\triangleright}}$ to stack these sparse input vectors, of which each row is an individual one-hot vector.

It is worth mentioning that the dynamic feature matrix $\textbf{G}^{\triangleright}$ is constructed in a chronological order. That is to say, $\textbf{G}^{\triangleright}$ can be viewed as a sequence of dynamic features, so for row $i < j$, $\textbf{g}^{\triangleright}_i \in \textbf{G}^{\triangleright}$ is always observed earlier than $\textbf{g}^{\triangleright}_j \!\in\! \textbf{G}^{\triangleright}$. As dynamic features may update frequently over time, we pose a threshold on the maximum sequence length that our model handles. To make the notations clear, we keep using $n^{\triangleright}$ to denote the maximum length for the dynamic feature sequence. If the dynamic feature sequence length is greater than the specified $n^{\triangleright}$, we consider the most recent $n^{\triangleright}$ features. If the sequence length is less than $n^{\triangleright}$, we repeatedly add a padding vector $\{0\}^{1\times m^{\triangleright}}$ to the top of $\textbf{G}^{\triangleright}$ until the length is $n^{\triangleright}$.

So far, we can rewrite the SeqFM model in Eq.(\ref{eq:SeqFM_OG}) as:
\begin{equation}\label{eq:SeqFM}
	\widehat{y} = w_0 + [(\textbf{G}^{\circ}\textbf{w}^{\circ})^{\top}; (\textbf{G}^{\triangleright}\textbf{w}^{\triangleright})^{\top}]\textbf{1} + f(\textbf{G}^{\circ}, \textbf{G}^{\triangleright}),
\end{equation}
where $\textbf{w}^{\circ} \!\! \in \! \mathbb{R}^{m^{\circ} \times 1}$ and $\textbf{w}^{\triangleright} \!\! \in \! \mathbb{R}^{m^{\triangleright} \times 1}$ are column vectors representing  weights for all features, $[\cdot;\cdot]$ denotes the horizontal concatenation of two vectors, and $\textbf{1}$ is a $(n^{\circ} \!+\! n^{\triangleright}) \times 1$ vector consisting of $1$s. In Eq.(\ref{eq:SeqFM}), the first two terms serve the same purpose as those in Eq.(\ref{eq:SeqFM_OG}), while $f(\textbf{G}^{\circ}, \textbf{G}^{\triangleright})$ denotes the multi-view self-attentive factorization scheme. The work flow of SeqFM is demonstrated in Figure \ref{Figure:model}. In what follows, we will describe the design of $f(\textbf{G}^{\circ}, \textbf{G}^{\triangleright})$ in detail. 

\subsection{Embedding Layer}
\begin{figure}[!t]
\center
\includegraphics[width = 3in]{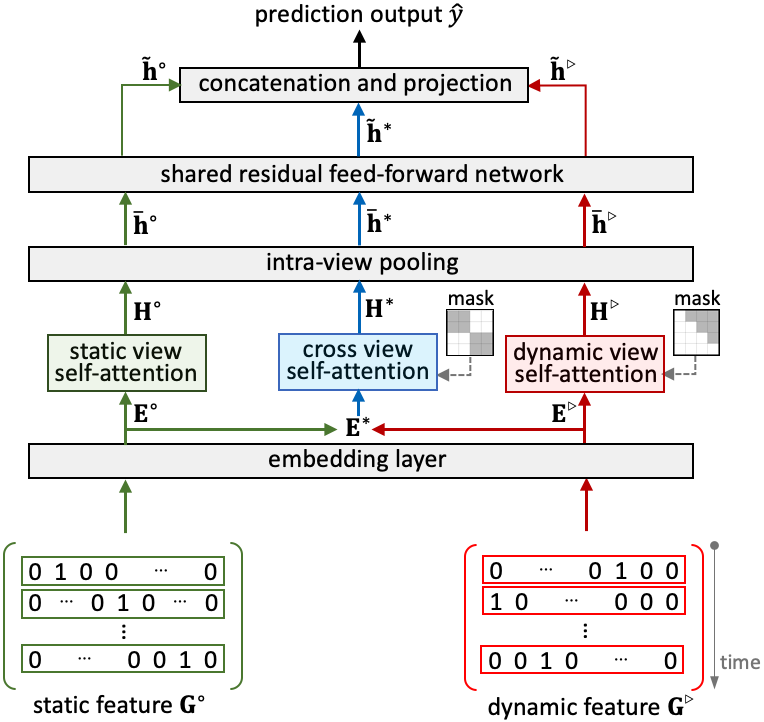}
\vspace{-0.4cm}
\caption{The overall architecture of SeqFM. We skip the linear term of SeqFM for better readability.}
\label{Figure:model}
\vspace{-0.5cm}
\end{figure}

As demonstrated in Figure \ref{Figure:model}, we first convert the sparse features $\textbf{G}^{\circ}$ and $\textbf{G}^{\triangleright}$ into dense representations with embedding. The embedding scheme is essentially a fully connected layer that projects each one-hot feature $\textbf{g}$ to a dense embedding vector as the following:
\vspace{-0.1cm}
\begin{equation}
\textbf{E}^{\circ} = \textbf{G}^{\circ} \textbf{M}^{\circ}, \hspace{0.1cm}
\textbf{E}^{\triangleright} = \textbf{G}^{\triangleright} \textbf{M}^{\triangleright},
\vspace{-0.1cm}
\end{equation}

where $\textbf{M}^{\circ} \!\in\! \mathbb{R}^{m^{\circ}\times d}$ and $\textbf{M}^{\triangleright} \!\in\! \mathbb{R}^{m^{\triangleright}\times d}$ are embedding matrices in the static and dynamic view, and $d$ is the latent embedding dimension. As such, we can obtain two embedded feature matrices $\textbf{E}^{\circ} \in \mathbb R^{n^{\circ} \times d}$ and $\textbf{E}^{\triangleright} \in \mathbb R^{n^{\triangleright} \times d}$, where each row is a embedding vector for the original feature. 

\subsection{Static View with Self-Attention}
From Eq.(\ref{eq:FM}), it is clear that in the traditional FM, feature interactions are modelled in a vector-wise manner \cite{lian2018xdeepfm}, where the dot product of two vectors is used.
To better encode the subtle and fine-grained information, recent FM-based models \cite{he2017neural,xiao2017attentional,qu2016product} shift to bit-wise (a.k.a. element-wise) interactions of feature embeddings, such as element-wise product and weighted sum. 
In order to comprehensively capture the complex interactions among features, we propose to jointly investigate vector-wise and bit-wise feature interactions with the self-attention \cite{vaswani2017attention}, which is a linear module that can be efficiently computed. We start with the self-attention module in the static view:
\vspace{-0.1cm}
\begin{equation}\label{eq:self_att}
	\textbf{H}^{\circ} = \textnormal{softmax} \Big{(}\frac{\textbf{Q}^{\circ} \textbf{K}^{\circ\top}}{\sqrt{d}}\Big{)}\textbf{V}^{\circ},
\vspace{-0.1cm}
\end{equation}
where $\textbf{H}^{\circ} \in \mathbb{R}^{n^{\circ} \times d}$ is the latent interaction representation for all $n^{\circ}$ static features, while $\sqrt{d}$ is the scaling factor to smooth the row-wise $SoftMax$ output and avoid extremely large values of the inner product, especially when the dimensionality is high. $\textbf{Q}^{\circ}$, $\textbf{K}^{\circ}$, $\textbf{V}^{\circ} \in \mathbb{R}^{n^{\circ} \times d}$ respectively represent the queries, keys and values obtained using linear projection:
\begin{equation}\label{eq:qkv}
	\textbf{Q}^{\circ} = \textbf{E}^{\circ} \textbf{W}_Q^{\circ}, \hspace{0.1cm}
	\textbf{K}^{\circ} = \textbf{E}^{\circ} \textbf{W}_K^{\circ}, \hspace{0.1cm}
	\textbf{V}^{\circ} = \textbf{E}^{\circ} \textbf{W}_V^{\circ},
\end{equation}
and $\textbf{W}_Q^{\circ}$, $\textbf{W}_K^{\circ}$, $\textbf{W}_V^{\circ} \in \mathbb{R}^{d \times d}$ are corresponding trainable projection weight matrices for queries, keys and values. 
To be concise, we reformulate the self-attention module in Eq.(\ref{eq:self_att}) and Eq.(\ref{eq:qkv}) as the following:
\begin{equation}\label{eq:static_att}
	\textbf{H}^{\circ} = \textnormal{softmax} \Big{(}\frac{\textbf{E}^{\circ} \textbf{W}_Q^{\circ} \cdot (\textbf{E}^{\circ}\textbf{W}_K^{\circ})^{\top}}{\sqrt{d}}\Big{)} \cdot \textbf{E}^{\circ}\textbf{W}_V^{\circ},
\end{equation}
and each row $\textbf{h}^{\circ}_i \!\in\! \textbf{H}^{\circ}$ corresponds to the $i$-th feature. Intuitively, we have $\textbf{h}^{\circ}_i \!=\! w_{i1}\textbf{v}^{\circ}_1 \!+\! w_{i2}\textbf{v}^{\circ}_2 \!+\! \cdots \!+\! w_{in^{\circ}}\textbf{v}^{\circ}_{n^{\circ}}$ where $w_{i1}, w_{i2},..., w_{in^{\circ}}$ are self-attentive weights assigned to projected features
$\textbf{v}^{\circ}_1, \textbf{v}^{\circ}_2,.., \textbf{v}^{\circ}_{n^{\circ}} \!\in\! \textbf{V}^{\circ}$.
%and it encodes the interactions between the $i$-th feature and all other features. 
In fact, because the vector-wise interactions are encoded via the self-attentive weights from the dot product scheme with $SoftMax$ normalization, and the bit-wise interactions are encoded in an additive form of features, the self-attention is able to account for both bit-wise and vector-wise feature interactions between the $i$-th feature and all other features. Furthermore, being able to learn asymmetric interactions, the projection operation with three distinctive subspaces makes the model more flexible.

\subsection{Dynamic View with Self-Attention}\label{sec:dvah}
In the dynamic view, due to the nature of sequential dependencies among $n^{\triangleright}$ dynamic features, the $i$-th dynamic feature ($i \leq n^{\triangleright}$) will only have the interactive influence from the previous features at $j$ ($j \leq i$). For example, in the movie rating prediction task, we can only infer a user's rating to a new movie from her/his ratings to the movies this user has watched before. That is to say, the feature interactions in the dynamic view are \textbf{\textit{directional}}. Thus, to incorporate the directional property into the self-attention mechanism, we propose the following masked self-attention to model the feature interactions in the dynamic view:  
\vspace{-0.1cm}
\begin{equation}\label{eq:self_att_dv}
	\textbf{H}^{\triangleright} = \textnormal{softmax} \Big{(}\frac{\textbf{E}^{\triangleright}\textbf{W}^{\triangleright}_Q \cdot (\textbf{E}^{\triangleright}\textbf{W}^{\triangleright}_K)^{\top}}{\sqrt{d}} + \textbf{M}^{\triangleright} \Big{)}\cdot \textbf{E}^{\triangleright}\textbf{W}^{\triangleright}_V,
\vspace{-0.2cm}
\end{equation}
where $\textbf{H}^{\triangleright} \in \mathbb{R}^{n^{\triangleright} \times d}$ carries the interaction contexts for all dynamic features, and $\textbf{W}_Q^{\triangleright}, \textbf{W}_K^{\triangleright}, \textbf{W}_V^{\triangleright} \in \mathbb{R}^{d \times d}$. Compared with other sequential approaches like recurrent neural networks, self-attention enables vector-wise feature interactions and is more computationally efficient \cite{kang2018self,vaswani2017attention}. Notably, $\textbf{M}^{\triangleright} \in \{-\infty,0\}^{n^{\triangleright} \times n^{\triangleright}}$ is a constant attention mask that allows each dynamic feature $\textbf{e}_{i}^{\triangleright}$ to interact with $\textbf{e}_{j}^{\triangleright}$ only if $j \!\leq\! i$. Specifically, in the mask $\textbf{M}^{\triangleright}$, for its row and column indexes $i,j \leq n^{\triangleright}$, the value of each entry $m^{\triangleright}_{ij} \in \textbf{M}^{\triangleright}$ is determined as:
\vspace{-0.2cm}
\begin{equation}\label{eq:mask_dv}
	m_{ij}^{\triangleright} = \Bigg{\{}
	\begin{array}{c}
			\hspace{0.1cm}0, \hspace{0.2cm} \textnormal{if}\hspace{0.2cm} i \geq j \\
			\hspace{-0.2cm}-\infty, \hspace{0.2cm} \textnormal{otherwise} \\
		\end{array}.\\
\vspace{-0.2cm}
\end{equation}

\textbf{The Rationale of Attention Mask.} 
We denote the matrix product of the query and key matrices in Eq.(\ref{eq:self_att_dv}) as $\textbf{A}$, i.e., $\textbf{A} \!\! = \!\! \frac{\textbf{E}^{\triangleright}\textbf{W}^{\triangleright}_Q \cdot (\textbf{E}^{\triangleright}\textbf{W}^{\triangleright}_K)^{\top}}{\sqrt{d}} \!\in\! \mathbb{R}^{n^{\triangleright} \times n^{\triangleright}}$. Similar to \cite{vaswani2017attention}, in $\textbf{A}$, each row $a_{i1}, a_{i2},..., a_{in^{\triangleright}}$ contains $n^{\triangleright}$ interaction scores between the $i$-th dynamic feature and all $n^{\triangleright}$ dynamic features. Then, for the $i$-th feature, $SoftMax$ is utilized to normalize these affinity scores to a probability distribution, i.e., $p_{i1}, p_{i2}, ..., p_{in^{\triangleright}} = \textnormal{softmax}(a_{i1}, a_{i2},..., a_{in^{\triangleright}})$. By adding the attention mask $\textbf{M}^{\triangleright}$, for the $i$-th feature, the interaction scores from $i+1$ become $-\infty$, while the earlier ones in the sequence remain unchanged. Consequently, with the $SoftMax$, $p_{ij} \!\! \neq \!\! 0$ for $j \!\leq\! i$ while $p_{ij} \!\approx\! 0$ for $j \!>\! i$, ensuring the interaction strength on the $i$-th feature only associates with historical features where $j \!\leq\! i$.

\subsection{Cross View with Self-Attention}
Because static and dynamic features possess varied semantics, in the cross view, we deploy the third attention head to model how static features interact with dynamic features. Similarly, we define another masked self-attention unit below:
%\vspace{-0.1cm}
\begin{equation}\label{eq:self_att_cv}
	\textbf{H}^{\star} = \textnormal{softmax} \Big{(}\frac{\textbf{E}^{\star}\textbf{W}^{\star}_Q \cdot (\textbf{E}^{\star}\textbf{W}^{\star}_K)^{\top}}{\sqrt{d}} + \textbf{M}^{\star} \Big{)}\cdot\textbf{E}^{\star}\textbf{W}_{V}^{\star},
\vspace{-0.1cm}
\end{equation}
where $\textbf{E}^{\star} \in \mathbb{R}^{(n^{\circ}+n^{\triangleright})\times d}$ represents the cross view feature matrix constructed by vertically concatenating feature matrices from both static and dynamic views along the first dimension:
\vspace{-0.2cm}
\begin{equation}
	\textbf{E}^{\star} \! = 
	\begin{bmatrix}
	\textbf{E}^{\circ} \\
	\textbf{E}^{\triangleright}\\
	\end{bmatrix}.
\vspace{-0.1cm}
\end{equation}

In Eq.(\ref{eq:self_att_cv}), $\textbf{H}^{\star} \in \mathbb{R}^{(n^{\circ}+n^{\triangleright})\times d}$ stacks the interaction contexts for all $n^{\circ}+n^{\triangleright}$ features, and there are corresponding query, key and value projection matrices $\textbf{W}^{\star}_Q, \textbf{W}^{\star}_K, \textbf{W}^{\star}_V \!\in\! \mathbb{R}^{d \times d}$. $\textbf{M}^{\star} \!\in\! \{-\infty,0\}^{(n^{\circ}+n^{\triangleright}) \times (n^{\circ}+n^{\triangleright})}$ is the attention mask devised for the cross view. Each entry $m_{ij}^{\star} \in \textbf{M}^{\star}$ is formulated via:
\vspace{-0.1cm}
\begin{equation}\label{eq:mask_cv}
	m_{ij}^{\star} = \Bigg{\{}
	\begin{array}{c}
			\hspace{0.3cm}0, \hspace{0.2cm} \textnormal{if}\hspace{0.2cm} i \leq n^{\circ} < j \hspace{0.2cm}\textnormal{or}\hspace{0.2cm} j \leq n^{\circ} < i \\
			\hspace{-3cm}-\infty, \hspace{0.2cm} \textnormal{otherwise} \\
		\end{array}.\\
\end{equation}

Following the explanation of the attention mask in Section~\ref{sec:dvah}, our cross view attention mask blocks possible feature interactions within the same category, and only allows cross-category feature interactions (i.e., interactions between static features and dynamic features). Intuitively, with this dedicated view, the model further extracts information from the mutual interactions between static properties and dynamic properties of features in a fine-grained manner.

\subsection{Intra-View Pooling Operation}
After calculating the representations for feature interactions in all three views, we feed these latent feature matrices into our proposed intra-view pooling layer, which compresses all latent vectors from each feature matrix into a unified vector representation. Specifically, for $\textbf{h}^{\circ}_i \!\in\! \textbf{H}^{\circ}$, $\textbf{h}^{\triangleright}_i \!\in\! \textbf{H}^{\triangleright}$ and $\textbf{h}^{\star}_i \!\in\! \textbf{H}^{\star}$, the intra-view pooling operation is defined as:
\vspace{-0.1cm}
\begin{equation}
\overline{\textbf{h}}^{view} = \frac{1}{n^{view}}\sum_{i=1}^{n^{view}}{\textbf{h}^{view}_i},
\vspace{-0.1cm}
\end{equation}
where $(\overline{\textbf{h}}^{view} , \textbf{h}^{view}_i , n^{view}) \in \{(\overline{\textbf{h}}^{\circ} , \textbf{h}^{\circ}_i , n^{\circ}), (\overline{\textbf{h}}^{\triangleright} , \textbf{h}^{\triangleright}_i , n^{\triangleright}),$ $ (\overline{\textbf{h}}^{\star} , \textbf{h}^{\star}_i , n^{\circ} + n^{\triangleright})\}$, and we use $\overline{\textbf{h}}^{\circ}$, $\overline{\textbf{h}}^{\triangleright}$ and $\overline{\textbf{h}}^{\star}$ to denote the final vector representations after the pooling operation for static view, dynamic view and cross view, respectively. 
Compared with the standard self-attention encoder structure in \cite{vaswani2017attention}, the intra-view pooling operation does not introduce additional model parameters. Moreover, the intra-view pooling operation compactly encodes the information of pairwise feature interactions in the static, dynamic and cross views.

\subsection{Shared Residual Feed-Forward Network}
With the multi-view self-attention and the intra-view pooling, all feature interactions are aggregated with adaptive weights. However, it is still a linear computation process. To further model the complex, non-linear interactions between different latent dimensions, we stack a shared $l$-layer residual feed-forward network upon the intra-view pooling layer:
\begin{equation}\label{eq:ffn}
\begin{split}
	&\widetilde{\textbf{h}}^{view}_{(1)} =  \overline{\textbf{h}}^{view} + \textnormal{ReLU}(\textnormal{LN}(\overline{\textbf{h}}^{view}) \textbf{W}_{1} + \textbf{b}_1),\\
	&\widetilde{\textbf{h}}^{view}_{(2)} =  \widetilde{\textbf{h}}^{view}_{(1)} + \textnormal{ReLU}(\textnormal{LN}(\widetilde{\textbf{h}}^{view}_{(1)}) \textbf{W}_2 + \textbf{b}_2),\\
	&\hspace{3.2cm} \small{\cdots}\\
	&\widetilde{\textbf{h}}^{view}_{(l)} = \widetilde{\textbf{h}}^{view}_{(l-1)} + \textnormal{ReLU}(\textnormal{LN}(\widetilde{\textbf{h}}^{view}_{(l-1)}) \textbf{W}_l + \textbf{b}_l),
\end{split}
\end{equation}
where $(\overline{\textbf{h}}^{view}, \widetilde{\textbf{h}}^{view}) \!\!\in\!\! \{(\overline{\textbf{h}}^{\circ}, \widetilde{\textbf{h}}^{\circ}), (\overline{\textbf{h}}^{\triangleright}, \widetilde{\textbf{h}}^{\triangleright}), (\overline{\textbf{h}}^{\star},\widetilde{\textbf{h}}^{\star})\}$, ReLU is the rectified linear unit for non-linear activation, while $\textbf{W}\in \mathbb{R}^{d\times d}$ and $\textbf{b} \in \mathbb{R}^{1\times d}$ are weight and bias in each layer. Note that though the network parameters are different from layer to layer, the three views share the same feed-forward network, as shown in Figure \ref{Figure:model}. In the following, we introduce the three key components in the shared residual feed-forward network.

\textbf{Residual Connections.}
The core idea behind residual networks is to propagate low-layer features to higher layers by residual connection \cite{he2016deep}. By combining low-layer interaction features with the high-layer representations computed by the feed-forward network, the residual connections essentially allow the model to easily propagate low-layer features to the final layer, which can help the model enhance its expressive capability using different information learned hierarchically. Intuitively, in our shared residual feed-forward network, to generate a comprehensive representation for feature interactions in each view, the $l$-th layer iteratively fine-tunes the representation learned by the $(l-1)$-th layer (i.e., $\widetilde{\textbf{h}}^{view}_{(l-1)}$) by adding a learned residual, which corresponds to the second term in Eq.(\ref{eq:ffn}).

\textbf{Layer Normalization.}
In Eq.(\ref{eq:ffn}), $\textnormal{LN}(\cdot)$ denotes the layer normalization function \cite{lei2016layer}, which is beneficial for stabilizing and accelerating neural network training process by normalizing the layer inputs across features. Unlike batch normalization \cite{ioffe2015batch}, in layer normalization, each sample from a batch uses independent statistics \cite{kang2018self}, and the computation at training and test times follows the same process. Specifically, for an arbitrary layer input $\widetilde{\textbf{h}}^{view}_{(l')},$ $\textnormal{LN}(\widetilde{\textbf{h}}^{view}_{(l')})$ is calculated as:
\begin{equation}\label{layer_norm}
	\textnormal{LN}(\widetilde{\textbf{h}}^{view}_{(l')}) = \textbf{s} \odot \frac{\widetilde{\textbf{h}}^{view}_{(l')} - \mu}{\epsilon} + \textbf{b},
\end{equation}
where $l' \leq l$ and $\widetilde{\textbf{h}}^{view}_{(l')} \in \{\widetilde{\textbf{h}}^{\circ}_{(l')}, \widetilde{\textbf{h}}^{\triangleright}_{(l')}, \widetilde{\textbf{h}}^{\star}_{(l')}\}$. Also, $\odot$ is the element-wise product, $\mu$ and $\epsilon$ are respectively the mean and variance of all elements in $\widetilde{\textbf{h}}^{view}_{(l')}$. Note that a small bias term will be added to $\epsilon$ in case $\epsilon\!=\!0$. The scaling weight $\textbf{s} \!\in\! \mathbb{R}^{1\times d}$ and the bias term $\textbf{b} \!\in\! \mathbb{R}^{1\times d}$ are parameters to be learned which help restore the representation power of the network.

\textbf{Layer Dropout.}
To prevent SeqFM from overfitting the training data, we adopt dropout \cite{srivastava2014dropout} on all the layers of our shared residual feed-forward network as a regularization strategy. In short, we randomly drop the neurons with the ratio of $\rho \in (0, 1)$ during training. Hence, dropout can be viewed as a form of ensemble learning which includes numerous models that share parameters \cite{warde2013empirical}. It is worth mentioning that all the neurons are used when testing, which can be seen as a model averaging operation \cite{srivastava2014dropout} in ensemble learning.

\subsection{View-Wise Aggregation}
With the $\widetilde{\textbf{h}}^{\circ}_{(l)}$, $\widetilde{\textbf{h}}^{\triangleright}_{(l)}$ and $\widetilde{\textbf{h}}^{\star}_{(l)}$ calculated by the $l$-layer shared residual feed-forward network, we perform view-wise aggregation to combine all the information from different types of feature interactions. The final representation is generated by horizontally concatenating the latent representations from three views:
\begin{equation}
	\textbf{h}^{agg} = [\widetilde{\textbf{h}}^{\circ}_{(l)}; \widetilde{\textbf{h}}^{\triangleright}_{(l)}; \widetilde{\textbf{h}}^{\star}_{(l)}],
\end{equation}
where $\textbf{h}^{agg} \in \mathbb{R}^{1\times 3d}$ denotes the aggregated representation of non-linear, high-order feature interactions within SeqFM. Since the representations learned by the shared residual feed-forward network are sufficiently expressive with an appropriate network depth $l$, we do not apply extra learnable weights to the view-wise aggregation scheme.

\subsection{Output Layer}
After the aggregation of the latent representations from the static, dynamic and cross views, the final vector representation $\textbf{h}^{agg}$ is utilized to compute the scalar output for the multi-view self-attentive factorization component via vector dot product:
\begin{equation}\label{eq:output0}
	f(\textbf{G}^{\circ}, \textbf{G}^{\triangleright}) = \langle \textbf{p}, \textbf{h}^{agg} \rangle,
\end{equation}
where $\textbf{p} \in \mathbb{R}^{1\times 3d}$ is the projection weight vector. At last, we summarize the entire prediction result of SeqFM as:
\begin{equation}\label{eq:output1}
	\widehat{y} = w_0 + [(\textbf{G}^{\circ}\textbf{w}^{\circ})^{\top}; (\textbf{G}^{\triangleright}\textbf{w}^{\triangleright})^{\top}]\textbf{1} + \langle\textbf{p},\textbf{h}^{agg}\rangle.
\end{equation}

As the scopes of both the input and output are not restricted, SeqFM is a flexible and versatile model which can be adopted for different tasks. In Section \ref{sec:app}, we will introduce how SeqFM is applied to ranking, classification, and regression tasks as well as the optimization strategy of SeqFM.

\subsection{Time Complexity Analysis}
Excluding the embedding operation that is standard in all FM-based models, the computational cost of our model is mainly exerted by the self-attention units and the feed-forward network. As the three self-attention units are deployed in parallel, we only consider the cross view attention head that takes the most time to compute. Hence, for each training sample, the overall time complexity of these two components is $O((n^{\circ}+n^{\triangleright})^{2}d) + O(ld^2) = O((n^{\circ}+n^{\triangleright})^2d+ld^2)$. Because $l$ is typically small, the dominating part is $O((n^{\circ}+n^{\triangleright})^2d)$. As $n^{\circ}$ is constant in the static view and $n^{\triangleright}$ is fixed with a threshold, SeqFM has linear time complexity w.r.t. the scale of the data.

\section{Applications and Optimization of SeqFM}\label{sec:app}
We hereby apply SeqFM to three different temporal predictive analytic settings, involving ranking, classification, and regression tasks. We also describe our optimization strategy.

\subsection{SeqFM for Ranking}
We deploy SeqFM for next-POI (point-of-interest) recommendation, which is commonly formulated as a ranking task \cite{yang2017bridging,yin2016adapting,wang2018tpm}. For each user, next-POI recommendation aims to predict a personalized ranking on a set of POIs and return the top-$K$ POIs according to the predicted ranking. This is accomplished by estimating a ranking score for each given user-POI pair $(user, POI)$. For this ranking task, the input of SeqFM is formulated as follows: 
\begin{equation}
	\!\textbf{G}^{\circ} \!\!=\!\!\! 
	\begin{bmatrix}
    \textbf{g}^{\circ}_1 \\ 
    \textbf{g}^{\circ}_2 \\
    \vdots \\
    \textbf{g}^{\circ}_{n^{\circ}} \\
    \end{bmatrix}
    \begin{array}{@{}l@{}}
    \!\! \rightarrow \! \textnormal{\small user one-hot}\\
    \!\! \rightarrow \! \textnormal{\small candidate POI one-hot}\\
    \!\! \Bigg{\}}  \,\,
    \begin{array}{@{}l@{}}
    \textnormal{\small other static}\\
    \textnormal{\small features}\\
    \end{array}\\
    \end{array}\!,    
	\textbf{G}^{\triangleright} \!\!=\!\!\! 
	\begin{bmatrix}
    \textbf{g}^{\triangleright}_1 \\ 
    \textbf{g}^{\triangleright}_2 \\
    \vdots \\
    \textbf{g}^{\triangleright}_{n^{\triangleright}} \\
    \end{bmatrix} \!\!
    \! \left \} 
  \begin{array}{@{}l@{}}
    \\
    \\
    \\
    \\
    \end{array}\right. 
    \begin{array}{@{}l@{}}
    \!\!\textnormal{\small one-hot sequence}\\
    \!\!\textnormal{\small of visited POIs}\\
    \end{array}\!.
\end{equation} 

Note that other static features include the user/POI's side information (e.g., occupation, gender, etc.) and are optional subject to availability. We denote the $(user, POI)$ pair as $(u, v)$ to be concise. For each user $u$, we denote an observed user-POI interaction as a positive pair $(u,v^+)$. Correspondingly, a corrupted user-POI pair $(u, v^-)$ can be constructed, where $v^-$ is a POI that user $u$ has never visited. Thus, a training sample is defined as a triple $(u_i, v_j^+, v_k^-) \in \mathcal{S}$, and $\mathcal{S}$ denotes the set of all training samples. Following \cite{rendle2009bpr}, we leverage the Bayesian Personalized Ranking (BPR) loss to optimize SeqFM for the ranking task:
\begin{align}
	\mathcal{L}
	& = - \log \prod_{(u_i, v_j^+, v_k^-)\in \mathcal{S}} \sigma(\widehat{y}_{ij} - \widehat{y}_{ik}) \nonumber\\
	& = - \sum_{(u_i, v_j^+, v_k^-)\in \mathcal{S}}{\log \Big{(}\sigma (\widehat{y}_{ij} - \widehat{y}_{ik}) \Big{)}},
\end{align}
where $\sigma(\cdot)$ is the $Sigmoid$ function. We omit the regularization term for model parameters as the layer dropout scheme is already capable of preventing our model from overfitting. For each user $u_i$, $\widehat{y}_{ij}$ and $\widehat{y}_{ik}$ respectively denote the ranking score for item $v_j^+$ and item $v_k^-$. The rationale of the BPR loss is that, the ranking score for a POI visited by the user should always be higher than the ranking score for an unvisited one.

\subsection{SeqFM for Classification}
For classification task, we conduct click-through rate (CTR) prediction, which is also one of the most popular applications for FM-based models \cite{lian2018xdeepfm,juan2016field,shan2016deep,cheng2016wide,guo2017deepfm}. Given an arbitrary user and her/his previously visited links (e.g., web pages or advertisements), the target of CTR prediction is to predict whether this user will click through a given link or not. We formulate the input of SeqFM for this classification task as:
\begin{equation}
	\!\!\textbf{G}^{\circ} \!\!=\!\!\! 
	\begin{bmatrix}
    \textbf{g}^{\circ}_1 \\ 
    \textbf{g}^{\circ}_2 \\
    \vdots \\
    \textbf{g}^{\circ}_{n^{\circ}} \\
    \end{bmatrix}
    \begin{array}{@{}l@{}}
    \!\! \rightarrow \! \textnormal{\small user one-hot}\\
    \!\! \rightarrow \! \textnormal{\small candidate link one-hot}\\
    \!\! \Bigg{\}}  \,\,
    \begin{array}{@{}l@{}}
    \textnormal{\small other static}\\
    \textnormal{\small features}\\
    \end{array}\\
    \end{array}\!,    
	\textbf{G}^{\triangleright} \!\!=\!\!\! 
	\begin{bmatrix}
    \textbf{g}^{\triangleright}_1 \\ 
    \textbf{g}^{\triangleright}_2 \\
    \vdots \\
    \textbf{g}^{\triangleright}_{n^{\triangleright}} \\
    \end{bmatrix} \!\!
    \! \left \} 
  \begin{array}{@{}l@{}}
    \\
    \\
    \\
    \\
    \end{array}\right. 
    \begin{array}{@{}l@{}}
    \!\!\textnormal{\small one-hot sequence}\\
    \!\!\textnormal{\small of clicked links}\\
    \end{array}\!.
\end{equation}

To enable the capability of classification, a $Sigmoid$ operation is added to the output layer. To keep the notations clear, we re-formulate the $\widehat{y}$ in Eq.(\ref{eq:output1}) as:
\begin{equation}
	\widehat{y} = \sigma(  w_0 + [(\textbf{G}^{\circ}\textbf{w}^{\circ})^{\top}; (\textbf{G}^{\triangleright}\textbf{w}^{\triangleright})^{\top}]\textbf{1} + \langle\textbf{p},\textbf{h}^{agg}\rangle ),
\end{equation}
where $\sigma(\cdot)$ denotes the $Sigmoid$ function. Here, the $\widehat{y} \in (0,1)$ can be viewed as the possibility of observing a $(user, link)$ instance. By replacing $(user, link)$ with the notion $(u,v)$, we quantify the prediction error with log loss, which is a special case of the cross-entropy:
\begin{align}
	\mathcal{L} & = - \sum_{(u_i, v_j^+)\in \mathcal{S}^+}{\!\!\! \log\widehat{y}_{ij}}  - \sum_{(u_i, v_j^-)\in \mathcal{S}^-}{\!\!\! \log(1 - \widehat{y}_{ij})} \nonumber\\
	& = - \sum_{(u_i, v_j)\in \mathcal{S}}{\!\!\! \Big{(} y_{ij}\log\widehat{y}_{ij} + (1-y_{ij})\log(1 - \widehat{y}_{ij}) \Big{)}},
\end{align}
where $\mathcal{S} = \mathcal{S}^+ \cap \mathcal{S}^-$ is the set of labeled $(u, v)$ pairs. Since we only have positive labels of observed interactions denoted by $(u,v^+)\in \mathcal{S}^+$, we uniformly sample negative labels $(u,v^-)\in\mathcal{S}^-$ from the unobserved interactions during training and control the number of negative samples w.r.t. the size of the positive ones. 

\subsection{SeqFM for Regression}
Finally, we apply SeqFM to a regression task, namely rating prediction which is useful for mining users' preferences and personalities \cite{wu2017recurrent,rendle2011fast}. We use the same problem setting as \cite{rendle2011fast,rendle2010factorization}, that is, given a user and her/his rated items, we estimate this user's rating to a new target item. SeqFM takes the following as its input:
\begin{equation}
	\!\textbf{G}^{\circ} \!=\!\! 
	\begin{bmatrix}
    \textbf{g}^{\circ}_1 \\ 
    \textbf{g}^{\circ}_2 \\
    \vdots \\
    \textbf{g}^{\circ}_{n^{\circ}} \\
    \end{bmatrix}
    \begin{array}{@{}l@{}}
    \!\! \rightarrow \! \textnormal{\small user one-hot}\\
    \!\! \rightarrow \! \textnormal{\small target item one-hot}\\
    \!\! \Bigg{\}}  \,\,
    \begin{array}{@{}l@{}}
    \textnormal{\small other static}\\
    \textnormal{\small features}\\
    \end{array}\\
    \end{array}\!,    
	\textbf{G}^{\triangleright} \!=\!\! 
	\begin{bmatrix}
    \textbf{g}^{\triangleright}_1 \\ 
    \textbf{g}^{\triangleright}_2 \\
    \vdots \\
    \textbf{g}^{\triangleright}_{n^{\triangleright}} \\
    \end{bmatrix} \!\!
    \! \left \} 
  \begin{array}{@{}l@{}}
    \\
    \\
    \\
    \\
    \end{array}\right. 
    \begin{array}{@{}l@{}}
    \!\textnormal{\small one-hot sequence}\\
    \!\textnormal{\small of rated items}\\
    \end{array}.
\end{equation} 

We denote each $(user,item)$ pair as $(u, v)$. For each $(u_i, v_j)$, the emitted output $\widehat{y}_{ij}$ is a continuous variable that tries to match up with the ground truth rating $y_{ij}$. Thus, we can directly apply the squared error loss below:
\begin{equation}
	\mathcal{L} = \sum _{(u_i,v_j) \in \mathcal{S}}{(\widehat{y}_{ij} - y_{ij})^{2}},
\end{equation}
where $S$ denotes the training set. Note that sampling negative training cases is unnecessary in the conventional rating prediction task.

\subsection{Optimization Strategy}\label{sec:opt}
As SeqFM is built upon the deep neural network structure, we can efficiently apply Stochastic Gradient Decent (SGD) algorithms to learn the model parameters by minimizing each task-specific loss $\mathcal{L}$. Hence, we leverage a mini-batch SGD-based algorithm, namely Adam \cite{kingma2014adam} optimizer. For different tasks, we tune the hyperparameters using grid search. Specifically, the latent dimension (i.e., factorization factor) $d$ is searched in $\{8,16,32,64,128\}$; the depth of the shared residual feed-forward network $l$ is searched in $\{1,2,3,4,5\}$; the maximum sequence length $n^{\triangleright}$ is searched in $\{10,20,30,40,50\}$; and the dropout ratio $\rho$ is searched in $\{0.5,0.6,0.7,0.8,0.9\}$. We will further discuss the impact of these key hyperparameters to the prediction performance of SeqFM in Section \ref{sec:expana}. For ranking and classification tasks, we draw 5 negative samples for each positive label during training. In addition, we set the batch size to 512 according to device capacity and the learning rate to $1\!\times\!10^{-4}$. We iterate the whole training process until $\mathcal{L}$ converges.

\section{Experimental Settings}\label{sec:expset}
In this section, we outline the evaluation protocols for our proposed SeqFM\footnote{Public access to codes: \\ http://bit.ly/SeqFM or http://bit.ly/bitbucket-SeqFM}. 

\subsection{Datasets}
To validate the performance of SeqFM in terms of ranking, classification, and regression, for each task we consider two real-world datasets, whose properties are introduced below.
\begin{itemize}
	\item \textbf{Gowalla (Ranking):} This is a global POI check-in dataset\footnote{https://snap.stanford.edu/data/loc-gowalla.html} collected from February 2009 to October 2010. Each user's visited POIs are recorded with a timestamp.
	\item \textbf{Foursquare (Ranking):} This POI check-in dataset\footnote{https://sites.google.com/site/yangdingqi/home/foursquare-dataset} is generated world-wide from April 2012 to September 2013, containing users' visited POIs at different times.
	\item \textbf{Trivago (Classification):} This dataset is obtained from the ACM RecSys Challenge\footnote{http://www.recsyschallenge.com/2019/} in 2019. It is a web search dataset consisting of users' visiting (e.g., clicking) logs on different webpage links.
	\item \textbf{Taobao (Classification):} It is a subset of user shopping log data released by Alibaba\footnote{https://tianchi.aliyun.com/}. We extract and sort users' clicking behavior on product links chronologically.
	\item \textbf{Beauty (Regression):} A series of users' product ratings\footnote{http://snap.stanford.edu/data/amazon/productGraph/} are crawled from Amazon from May 1996 to July 2014, and different product categories are treated as separate datasets. Beauty is one of the largest categories.
	\item \textbf{Toys (Regression):} This is another Amazon user rating dataset on toys and games.
\end{itemize}

All datasets used in our experiment are in large scale and publicly available. The primary statistics are shown in Table~\ref{table:Dataset}, where we use the word ``object" to denote the POI, link, and item in different applications. Following \cite{yang2017bridging,li2016point,rendle2009bpr,rendle2010factorizing}, we filter out inactive users with less than 10 interacted objects and unpopular objects visited by less than 10 users. Note that for Beauty and Toys, we directly use the provided versions without further preprocessing.

%``\#" is the number of each object, while ``object" refers to POI, link, and item respectively in ranking, classification and regression datasets.
\begin{table}[!t]
\small
\caption{Statistics of datasets in use.}
\vspace{-0.8cm}
\renewcommand{\arraystretch}{0.55}
\setlength\tabcolsep{3.5pt}
\center
  \begin{tabular}{c c c c c c}
    \toprule
    \multirow{2}{*}{Task} & \multirow{2}{*}{Dataset} & \multirow{2}{*}{\#Instance} & \multirow{2}{*}{\#User} & \multirow{2}{*}{\#Object} & \#Feature\\
    &&&&&(Sparse)\\
	\midrule
	\multirow{2}{*}{Ranking} & Gowalla & 1,865,119 & 34,796 & 57,445 & 149,686\\
    		    & Foursquare & 1,196,248 & 24,941 & 28,593 & 82,127 \\
    \midrule
   \multirow{2}{*}{Classification} & Trivago & 2,810,584 & 12,790 & 45,195 & 103,180\\
    & Taobao & 1,970,133 & 37,398 & 65,474 & 168,346\\
    \midrule
    \multirow{2}{*}{Regression} & Beauty & 198,503 & 22,363 & 12,101 & 46,565\\
    & Toys & 167,597 & 19,412 & 11,924 & 50,748\\
    \bottomrule
\end{tabular}
\label{table:Dataset}
\vspace{-0.6cm}
\end{table}

\subsection{Baseline Methods}
We briefly introduce the baseline methods for comparison below. First of all, we choose the latest and popular FM-based models as the common baselines for all ranking, classification, and regression tasks. Then, for each task, we further select two state-of-the-art methods originally proposed for the specific task scenario as an additional competitor.
\begin{itemize}
	\item \textbf{FM:} This is the original Factorization Machine \cite{rendle2010factorization} with proven effectiveness in many prediction tasks \cite{rendle2011fast, rendle2012factorization}.
	\item \textbf{Wide\&Deep:} The Wide\&Deep \cite{cheng2016wide} model uses a DNN to learn latent representations of concatenated features. 
%(the deep part), which is coupled with a linear regression model (the wide part).
	\item \textbf{DeepCross:} It stacks multiple residual network blocks upon the concatenation layer for feature embeddings in order to learn deep cross features \cite{shan2016deep}. 
	\item \textbf{NFM:} The Neural Factorization Machine \cite{he2017neural} encodes all feature interactions via multi-layer DNNs coupled with a bit-wise bi-interaction pooling layer.
	\item \textbf{AFM:} The Attentional Factorization Machine \cite{xiao2017attentional}	  introduces an attention network to distinguish the importance of different pairwise feature interactions.
	%\item  in order to distinguish important feature interactions from the irrelevant ones.
	\item \textbf{SASRec (Ranking):} This is the Self-Attention-based Sequential Recommendation Model \cite{kang2018self} with long-term and short-term context modelling.
	\item \textbf{TFM (Ranking):} The Translation-based Factorization Machine \cite{pasricha2018translation} learns an embedding and translation space for each feature dimension, and adopts Euclidean distance to quantify the strength of pairwise feature interactions.
	% Designed for sequential recommendation, the 
	\item \textbf{DIN (Classification):} The Deep Interest Network \cite{zhou2018deep} can represent users' diverse interests with an attentive activation mechanism for CTR prediction.
	\item \textbf{xDeepFM (Classification):} It stands for the Extreme Deep Factorization Machine \cite{lian2018xdeepfm}, which has a compressed interaction network to model vector-wise feature interactions to perform CTR prediction.
\item \textbf{RRN (Regression):} The Recurrent Recommender Network \cite{wu2017recurrent} is a deep autoregressive model for temporal rating prediction.
\item \textbf{HOFM (Regression):} This is the Higher-Order Factorization Machine described in \cite{blondel2016higher}. HOFM improves \cite{rendle2010factorization} with space-saving and time-efficient kernels to allow shared parameters for prediction tasks.
\end{itemize}

\subsection{Evaluation Metrics}
To fit the scenario of temporal predictive analytics, we adopt the \textit{\textbf{leave-one-out}} evaluation protocol which is widely used in the literature \cite{he2017neuralcol,rendle2009bpr,zhou2018deep,pasricha2018translation}. Specifically, within each user's transaction, we hold out her/his last record as the ground truth for test and the second last record for validation. All the rest records are used to train the models. Set-category features are used as input for all FM-based baseline models.

\textbf{Evaluating Ranking Performance.}
To evaluate the ranking performance, we adopt the well-established Hits Ratio at Rank $K$ ($\textnormal{HR}@K$) and Normalized Discounted Cumulative Gain at Rank $K$ ($\textnormal{NDCG}@K$) which are commonly used in information retrieval and recommender systems \cite{yang2017bridging,chen2019air,yin2015modeling}. Specifically, for each positive test instance $(user, POI) \in \mathcal{S}^{test}$, we mix the POI with $J$ random POIs that are never visited by the user. Afterwards, we rank all these $J+1$ POIs for the user. Then, we use $\textnormal{HR}@K$ to measure the ratio that the ground truth item has a hit (i.e., is present) on the top-$K$ list, and use $\textnormal{NDCG}@K$ to further evaluate whether if the model can rank the ground truth as highly as possible:
\vspace{-0.1cm}
\begin{equation}\label{eq:hitsatk}
	\textnormal{HR}@K\!=\!\frac{\#hit@K}{|\mathcal{S}^{test}|}, \,\textnormal{NDCG}@K\!=\!\frac{\sum_{s\in \mathcal{S}^{test}} \! \sum_{r=1}^{K} \!\frac{rel_{s,r}}{\log_2(r+1)}}{|\mathcal{S}^{test}|},
\end{equation}
where $\#hit@K$ is the number of hits in the test set. For each test case $s \!\in\! \mathcal{S}^{test}$, $rel_{s,r} \!=\! 1$ if the item ranked at $r$ is the ground truth, otherwise $rel_{s,r} \!=\! 0$. We set $J\!=\!1,000$ to balance the running time and task difficulty. For $K$, we adopt the popular setting of $5,10,20$ for presentation.

\textbf{Evaluating Classification Performance.}
We adopt two evaluation metrics for the classification task, namely Area under the ROC Curve (AUC) \cite{lian2017practical,shan2016deep} and Root Mean Squared Error (RMSE) \cite{he2017neural,xiao2017attentional}. For each positive test instance $(user, link) \in \mathcal{S}^{test}$, we draw a random negative link that the user has never clicked, and predict the interaction possibility for both links. %AUC measures the probability that a positive instance will be ranked higher than the negative one. It only takes into account the order of predicted instances and is insensitive to class imbalance problem. In contrast, RMSE evaluates the distance between the predicted possibility and the true label for each instance.
 
%cheng2014gradient,
\textbf{Evaluating Regression Performance.}
We evaluate the regression performance with Mean Absolute Error (MAE) and Root Relative Squared Error (RRSE), which are popular among relevant research communities \cite{lai2017modeling,chen2018tada,chen2019online}. Mathematically, they are defined as follows:
\vspace{-0.2cm}
\begin{equation}
	\textnormal{MAE}=\frac{\sum_{y \in \mathcal{S}^{test}}|\widehat{y} - y|}{|\mathcal{S}^{test}|},
	\textnormal{RRSE}=\frac{\sqrt {\frac{\sum_{y \in |\mathcal{S}^{test}|}(\widehat{y} - y)^{2}}{|\mathcal{S}^{test}|}}}{VAR_{\mathcal{S}^{test}}},
\end{equation}
where $\widehat{y}$ and $y$ denote the predicted and real value respectively, and $VAR_{\mathcal{S}^{test}}$ is the variance of all ground truth values. %MAE directly reveals the gap between prediction and ground truth, while RRSE is the normalized root mean square error and is independent of the data scale and distribution.

\subsection{Parameter Settings}
To be consistent, we report the overall performance of SeqFM on all tasks with a unified parameter set $\{d=64, l=1, n^{\triangleright}=20, \rho=0.6\}$. Detailed discussions on the effects of different parameter settings will be shown in Section \ref{sec:paramterimpact}. For all baseline methods, since all tasks are conducted on standard and generic datasets with common evaluation metrics, we adopt the optimal parameters in their original works.

\begin{table*}[!t]
\small
\vspace{-0.2cm}
\caption{Ranking task (next-POI recommendation) results. Numbers in bold face are the best results for corresponding metrics.}
\vspace{-0.3cm}
\centering
\renewcommand{\arraystretch}{1.0}
\setlength\tabcolsep{7pt}
  \begin{tabular}{|c|c|c|c|c|c|c||c|c|c|c|c|c|}
    \hline
    \multirow{3}{*}{Method} & \multicolumn{6}{c||}{Gowalla}  & \multicolumn{6}{c|}{Foursquare}\\
    \cline{2-13}
    & \multicolumn{3}{c|}{HR$@$K} & \multicolumn{3}{c||}{NDCG$@$K} & \multicolumn{3}{c|}{HR$@$K} & \multicolumn{3}{c|}{NDCG$@$K}\\
    \cline{2-13}
    & K=5& K=10 & K=20 & K=5 & K=10 & K=20 & K=5& K=10 & K=20 & K=5 & K=10 & K=20\\
    \hline
    FM \cite{rendle2010factorization}& 0.232 & 0.318 & 0.419 & 0.158 & 0.187 & 0.211 & 0.241 & 0.303 & 0.433 & 0.169 & 0.201 & 0.217 \\
    Wide\&Deep \cite{cheng2016wide}& 0.288 & 0.401 & 0.532 & 0.199 & 0.238 & 0.267 & 0.233 & 0.317 & 0.422 & 0.165 & 0.192 & 0.218 \\
    DeepCross \cite{shan2016deep}& 0.273 & 0.379 & 0.505 & 0.182 & 0.204 & 0.241 & 0.282 & 0.355 & 0.492 & 0.198 & 0.210 & 0.229 \\
    NFM\cite{he2017neural}& 0.286 & 0.395 & 0.525 & 0.199 & 0.236 & 0.264 & 0.239 & 0.325 & 0.435 & 0.170 & 0.198 & 0.225 \\
    AFM\cite{xiao2017attentional}& 0.295 & 0.407 & 0.534 & 0.204 & 0.242 & 0.270 & 0.279 & 0.379 & 0.504 & 0.199 & 0.212 & 0.233 \\
    SASRec\cite{kang2018self}& 0.310 & 0.424 & 0.559 & 0.209 & 0.253 & 0.285 & 0.266 & 0.350 & 0.467 & 0.175 & 0.204 & 0.216\\    
    TFM\cite{pasricha2018translation}& 0.307 & 0.430 & 0.556 & 0.216 & 0.256 & 0.283 & 0.283 & 0.390 & 0.512 & 0.203 & 0.223 & 0.248\\
    \hline
     \textbf{SeqFM}& \textbf{0.345} & \textbf{0.467} & \textbf{0.603} & \textbf{0.243} & \textbf{0.283} & \textbf{0.316} & \textbf{0.324} & \textbf{0.431} & \textbf{0.554} & \textbf{0.227} & \textbf{0.262} & \textbf{0.293} \\
     \hline
    \end{tabular}
\label{table:ranking}
\vspace{-0.8cm}
\end{table*}

\begin{table}[t]
\small
\caption{Classification task (CTR prediction) results. Numbers in bold face are the best results for corresponding metrics.}
\vspace{-0.3cm}
\centering
\renewcommand{\arraystretch}{1.0}
\setlength\tabcolsep{7pt}
  \begin{tabular}{|c|c|c||c|c|}
    \hline
    \multirow{2}{*}{Method} & \multicolumn{2}{c||}{Trivago}  & \multicolumn{2}{c|}{Taobao}\\
    \cline{2-5}
    & AUC & RMSE & AUC & RMSE\\
    \hline
    FM \cite{rendle2010factorization}& 0.729 & 0.564 & 0.602 & 0.597\\
    Wide\&Deep \cite{cheng2016wide}& 0.782 & 0.529 & 0.629 & 0.590  \\
    DeepCross \cite{shan2016deep}& 0.845 & 0.433 & 0.735 & 0.391 \\
    NFM\cite{he2017neural}& 0.767 & 0.537 & 0.616 & 0.583 \\
    AFM\cite{xiao2017attentional}&  0.811 & 0.465 & 0.656 & 0.544\\
    DIN\cite{zhou2018deep}& 0.923 & 0.338 & 0.781 & 0.375\\
    xDeepFM\cite{lian2018xdeepfm}& 0.913 & 0.350 & 0.804 & 0.363\\
    \hline
     \textbf{SeqFM}& \textbf{0.957} & \textbf{0.319} & \textbf{0.826} & \textbf{0.335}\\
     \hline
    \end{tabular}
\label{table:classification}
\vspace{-0.7cm}
\end{table}

\section{Experimental Results and Analysis}\label{sec:expana}
Following the settings in Section \ref{sec:expset}, we conduct experiments to showcase the advantage of SeqFM in terms of both effectiveness and efficiency. In particular, we aim to answer the following research questions (RQs) via experiments:
\begin{description}
	\item[\textbf{RQ1:}] How effectively SeqFM can perform temporal predictive analytics compared with state-of-the-art FM-based models.
	\item[\textbf{RQ2:}] How the hyperparameters affect the performance of SeqFM in different prediction tasks.
	\item[\textbf{RQ3:}] How SeqFM benefits from each component of the proposed model structure.
	\item[\textbf{RQ4:}] How is the training efficiency and scalability of SeqFM when handling large-scale data. 
\end{description}

\subsection{Prediction Performance (RQ1)}
We summarize the performance of all models in terms of ranking, classification, and regression with Table \ref{table:ranking}, \ref{table:classification}, and \ref{table:regression} respectively. We discuss our findings as follows.

\textbf{Ranking Performance.}\label{sec:rankperform}
The results of the ranking task (next-POI recommendation) are reported in Table \ref{table:ranking}. Note that higher HR$@K$ and NDCG$@K$ values imply better prediction performance. Obviously, on both Gowalla and Foursquare, SeqFM significantly and consistently outperforms all existing FM-based models with $K \in \{5,10,20\}$. In particular, the advantages of SeqFM in terms of HR$@$5 and NDCG$@$5 imply that our model can accurately rank the ground truth POI in the top-5 positions, which can better suit each user's intent and boost the recommendation success rate. Though SASRec shows promising effectiveness on Gowalla, it underperforms when facing higher data sparsity on Foursquare. Another observation is that all FM-based models with deep neural networks (i.e., Wide\&Deep, DeepCross, NFM and AFM) outperform the plain FM. As a model specifically designed for sequential recommendation, TFM naturally performs better than the common baselines on both POI check-in datasets. However, SeqFM still achieves higher ranking effectiveness. This is because TFM is designed to only consider the most recently visited object (POI) in the dynamic feature sequence, while SeqFM utilizes the self-attention mechanism to extract richer information from the entire sequence.

\textbf{Classification Performance.} We list all the results of the classification task (CTR prediction) in Table \ref{table:classification}. A better result corresponds to a higher AUC score and a lower RMSE value. At the first glance, it is clear that our SeqFM achieves the highest classification accuracy on both Trivago and Taobao. Similar to the observations from the ranking task, exisiting variants of the plain FM show the benefit of adopting deep neural networks. As for the task-specific models for CTR prediction, the attentive activation unit in DIN can selectively determine the weights of different features based on a given link, while xDeepFM is able to thoroughly model the high-order interactions among different features with its dedicated interaction network. However, there is a noticeable performance gap between both additional baselines and our proposed SeqFM. This proves the insight of our work, which points out that instead of simply treating all dynamic features as flat set-category features in existing FM-based models, the sequence-aware interaction scheme for dynamic features in SeqFM is more helpful for temporal predictive analytics.

\begin{table}[t!]
\small
\caption{Regression task (rating prediction) results. Numbers in bold face are the best results for corresponding metrics.}
\vspace{-0.3cm}
\centering
\renewcommand{\arraystretch}{1.0}
\setlength\tabcolsep{7pt}
  \begin{tabular}{|c|c|c||c|c|}
    \hline
    \multirow{2}{*}{Method} & \multicolumn{2}{c||}{Beauty}  & \multicolumn{2}{c|}{Toys}\\
    \cline{2-5}
    & MAE & RRSE & MAE & RRSE\\
    \hline
    FM \cite{rendle2010factorization}& 1.067 & 1.125 & 0.778 & 1.023\\
    Wide\&Deep \cite{cheng2016wide}& 0.965 & 1.090 & 0.753 & 0.989\\
    DeepCross \cite{shan2016deep}& 0.949 & 1.003 & 0.761 & 1.010\\
    NFM\cite{he2017neural}& 0.931 & 0.986 & 0.735 & 0.981\\
    AFM\cite{xiao2017attentional}& 0.945 & 0.994 & 0.741 & 0.997\\
    RRN\cite{wu2017recurrent} & 0.943 & 0.989 & 0.739 & 0.983\\
    HOFM\cite{blondel2016higher}& 0.952 & 1.054 & 0.748 & 1.001\\
    \hline
     \textbf{SeqFM}& \textbf{0.890} & \textbf{0.975} & \textbf{0.704} & \textbf{0.956}\\
     \hline
    \end{tabular}
\label{table:regression}
\vspace{-0.6cm}
\end{table}

\textbf{Regression Performance.} Table \ref{table:regression} reveals all models' performance achieved in the regression task (rating prediction) on Beauty and Toys. For both MAE and RRSE metrics, the lower the better. As demonstrated by the results, despite the intense competition in the regression task, SeqFM yields significant improvements on the regression accuracy over all the baselines. Furthermore, though showing competitive regression results, the additional baseline HOFM is still limited by its linear mathematical form, so approaches based on deep neural networks like RRN, NFM and AFM perform slightly better owing to their non-linear expressiveness. Apart from that, we notice that compared with the performance achieved by the plain FM, other FM-based approaches only shows marginal advantages against it in the regression task. In contrast, with 13\% and 7\% relative improvements on RRSE over the plain FM, our proposed SeqFM highlights the importance of fully utilizing the sequential dependencies for predictive analytics. 

To summarize, the promising effectiveness of SeqFM is thoroughly demonstrated in ranking, classification, and regression tasks. In the comparison with state-of-the-art baselines on a wide range of datasets, the considerable improvements from our model further imply that SeqFM is a general and versatile model that suits different types of temporal prediction tasks.

\begin{figure*}[t!]
\centering
\vspace{-0.4cm}
\begin{tabular}{cccc}
	\vspace{-0.2cm}\includegraphics[width=1.8in]{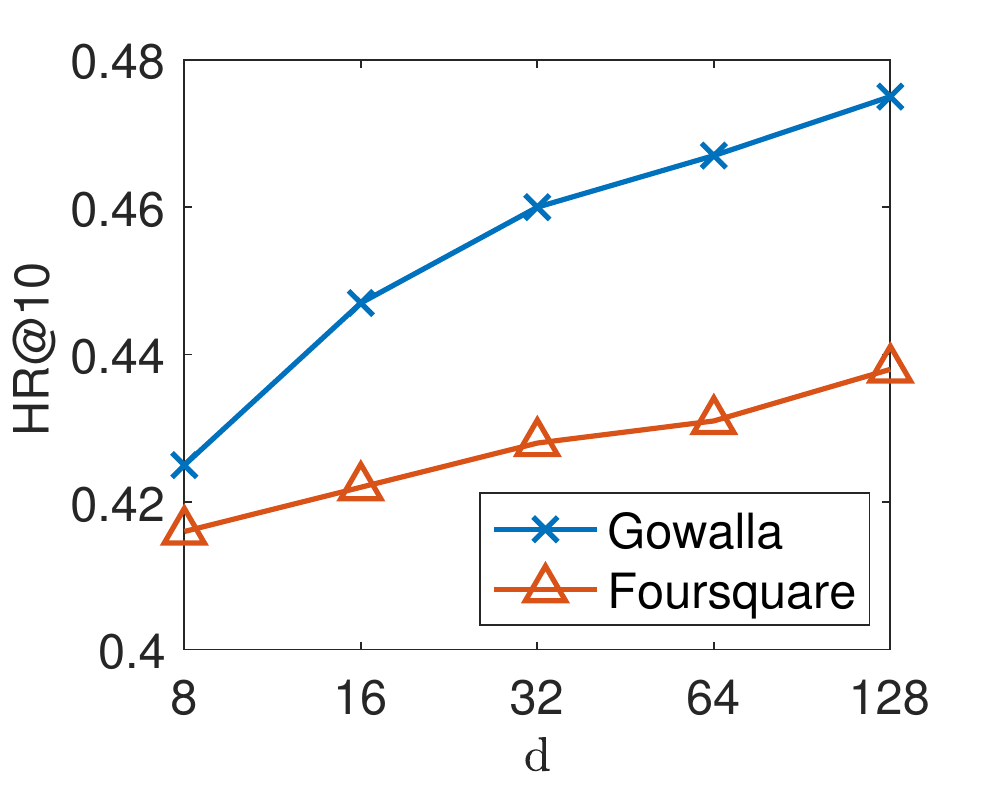}
	&\hspace{-0.6cm}\includegraphics[width=1.8in]{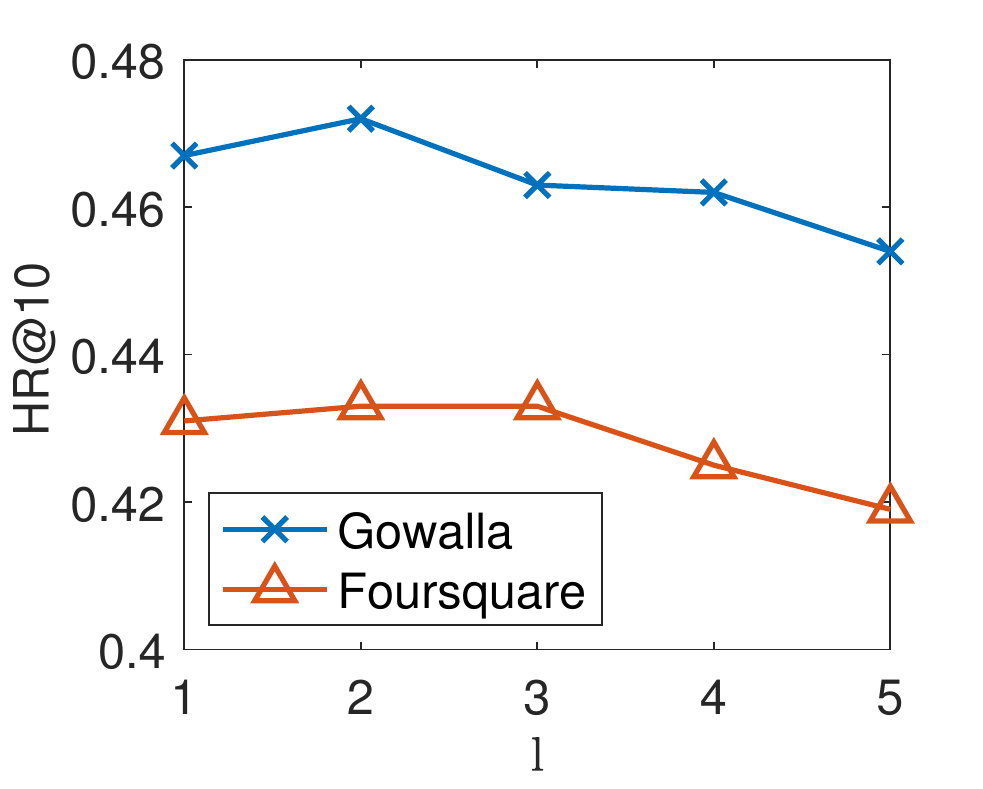}
	&\hspace{-0.6cm}\includegraphics[width=1.8in]{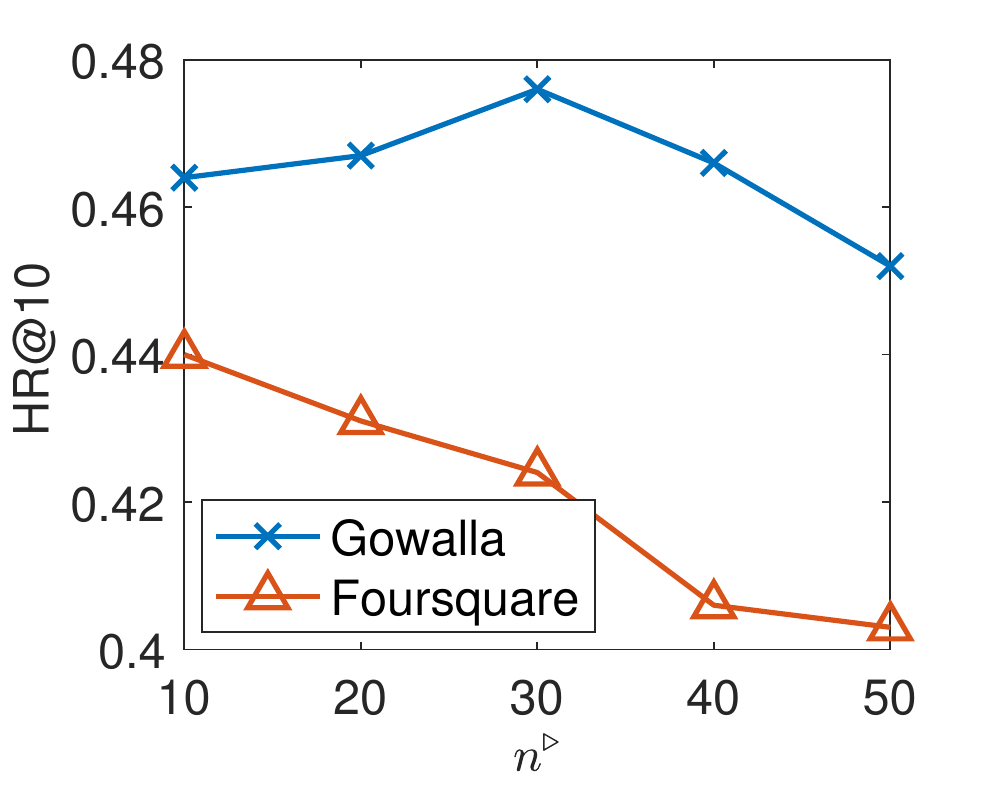}
	&\hspace{-0.6cm}\includegraphics[width=1.8in]{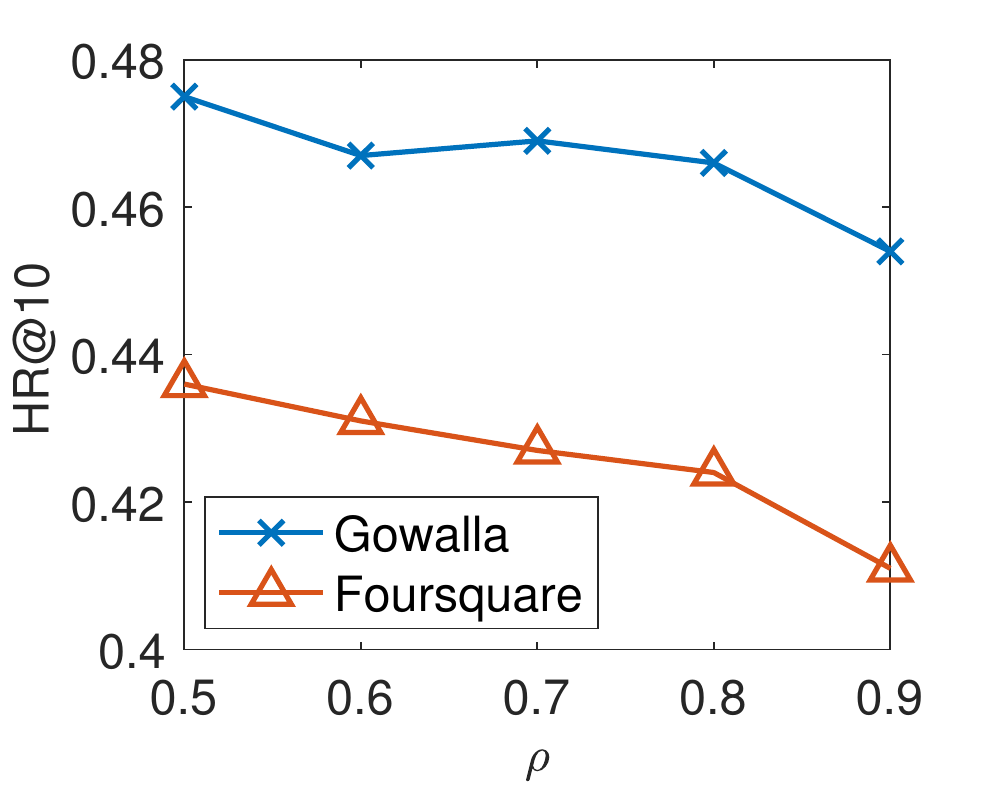}\\
	 \multicolumn{4}{c}{\hspace{0.5cm}\footnotesize (a) Above: ranking performance w.r.t. $d$, $l$, $n^{\triangleright}$ and $\rho$.} \\
		\vspace{-0.2cm}\includegraphics[width=1.8in]{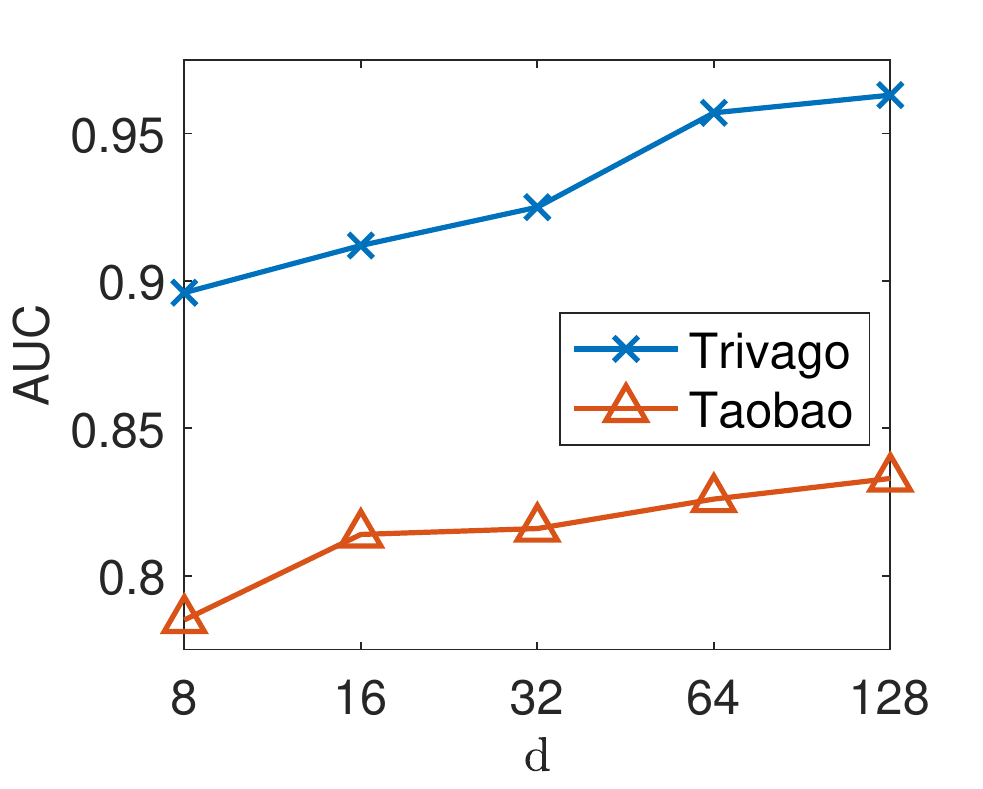}
	&\hspace{-0.6cm}\includegraphics[width=1.8in]{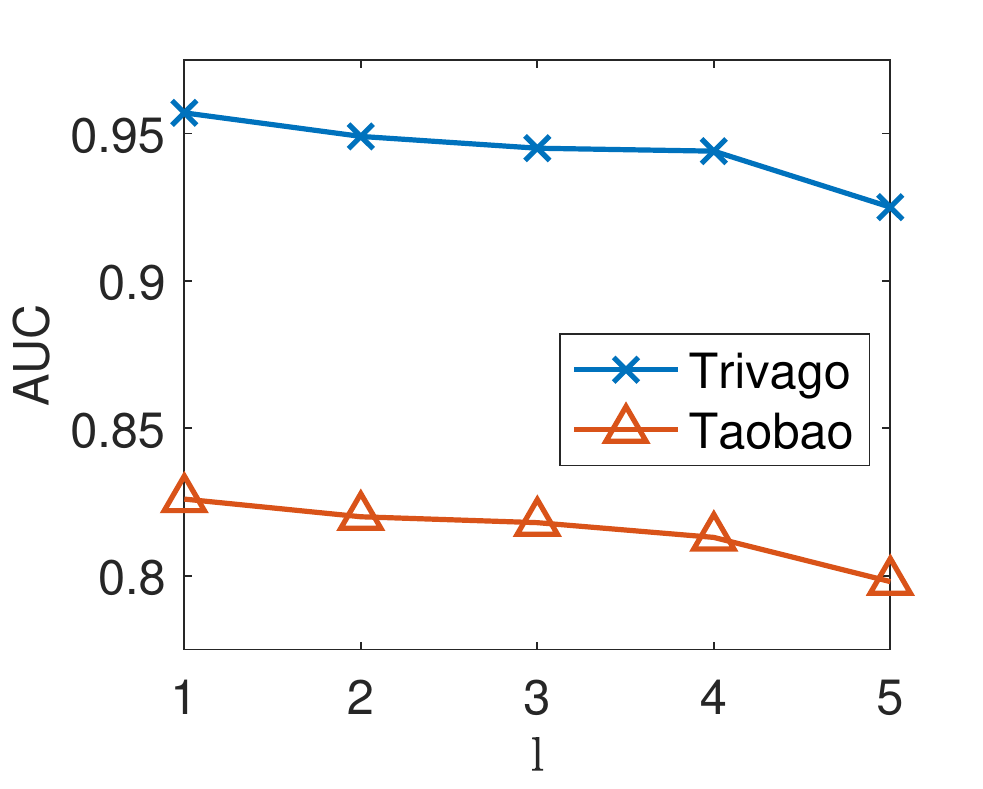}
	&\hspace{-0.6cm}\includegraphics[width=1.8in]{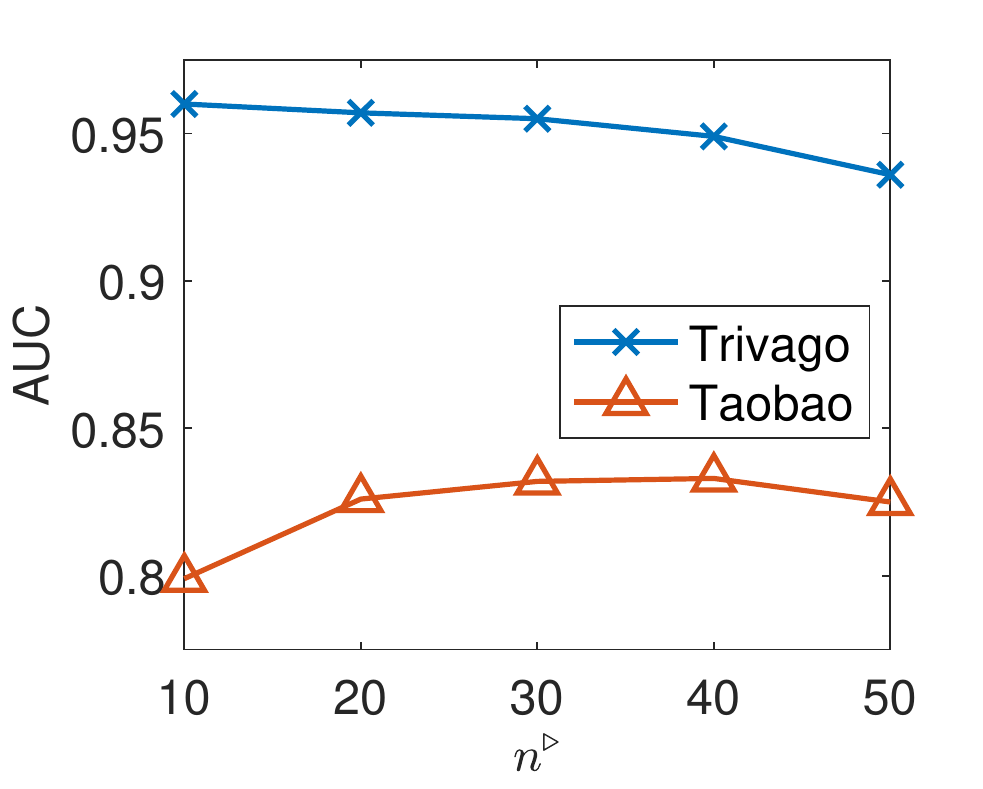}
	&\hspace{-0.6cm}\includegraphics[width=1.8in]{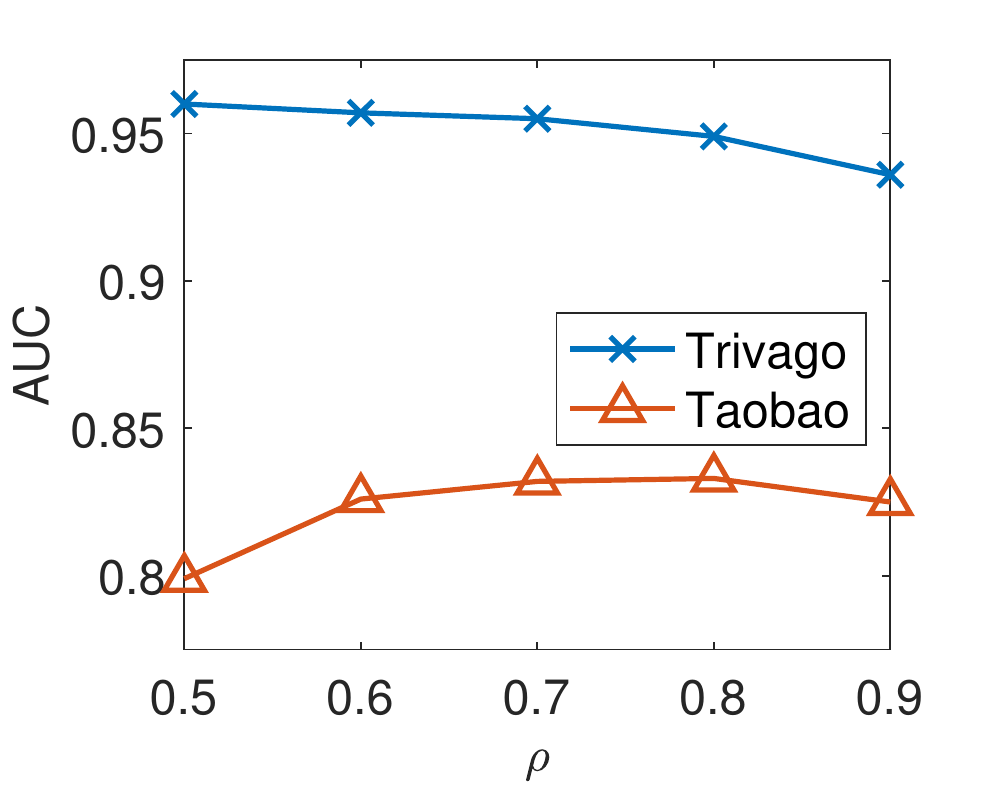}\\
	\multicolumn{4}{c}{\hspace{0.5cm}\footnotesize (b) Above: classification performance w.r.t. $d$, $l$, $n^{\triangleright}$ and $\rho$.} \\
		\vspace{-0.2cm}\includegraphics[width=1.8in]{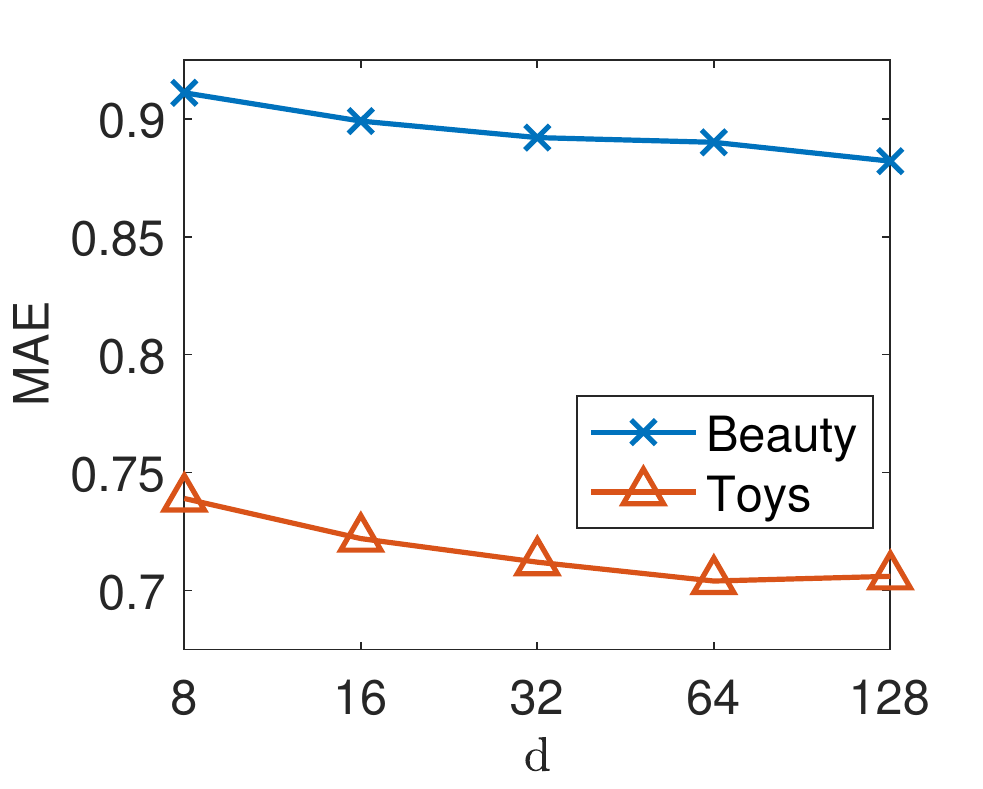}
	&\hspace{-0.6cm}\includegraphics[width=1.8in]{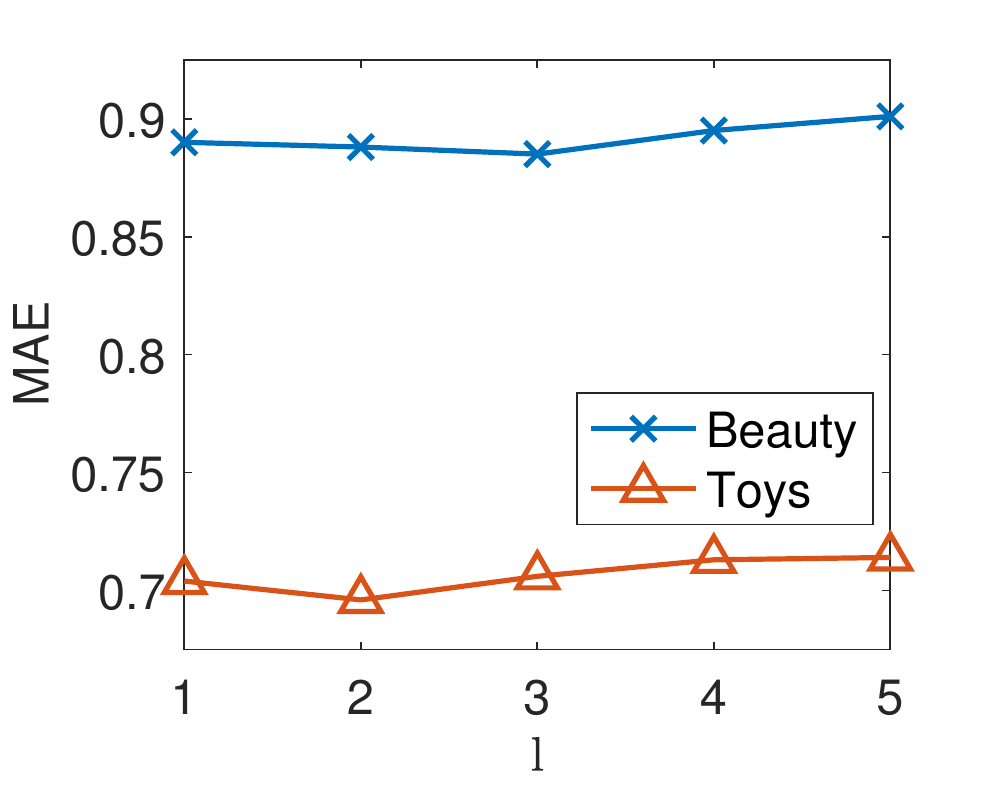}
	&\hspace{-0.6cm}\includegraphics[width=1.8in]{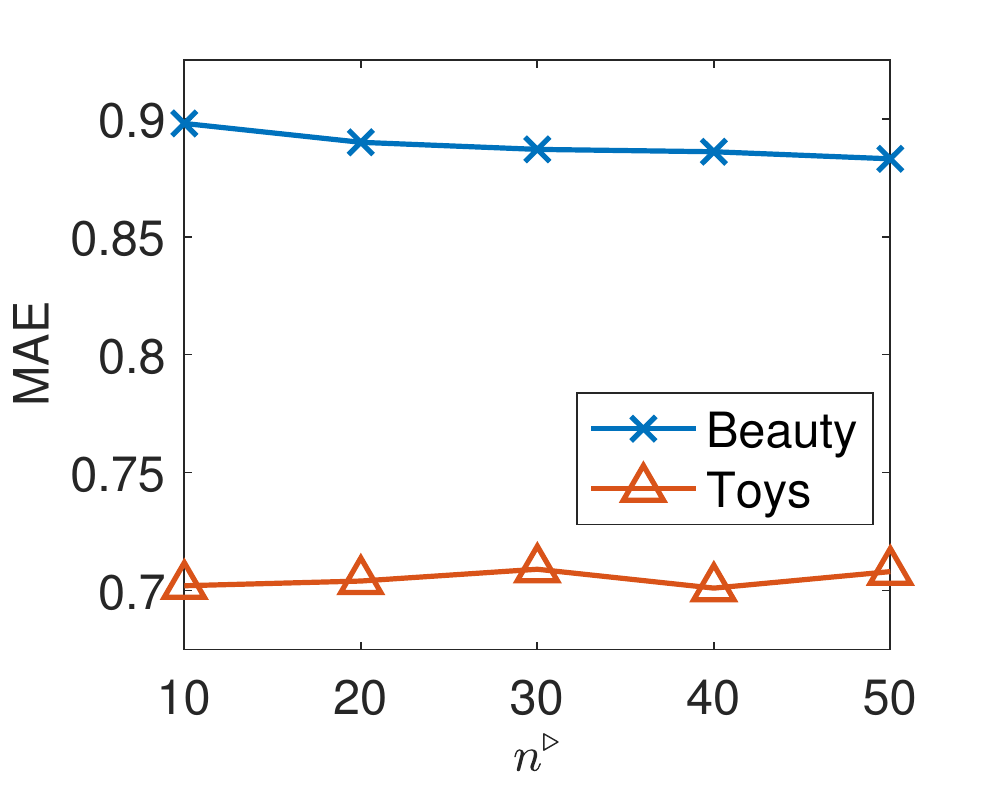}
	&\hspace{-0.6cm}\includegraphics[width=1.8in]{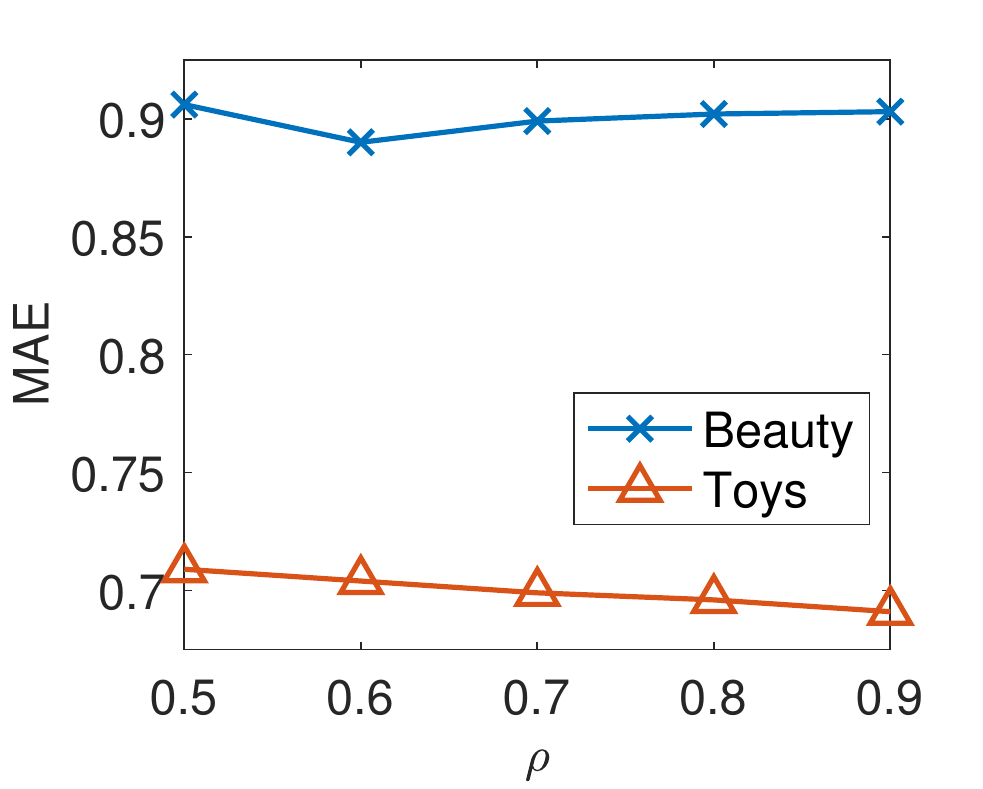}\\
	\multicolumn{4}{c}{\hspace{0.5cm}\footnotesize (c) Above: regression performance w.r.t. $d$, $l$, $n^{\triangleright}$ and $\rho$.}\\
\end{tabular}
\vspace{-0.4cm}
\caption{Parameter sensitivity analysis.}
\label{Figure:paramsensitivity}
\vspace{-0.7cm}
\end{figure*}

\subsection{Impact of Hyperparameters (RQ2)}\label{sec:paramterimpact}
We answer the second research question by investigating the performance fluctuations of SeqFM with varied hyperparameters. Particularly, as mentioned in Section \ref{sec:opt}, we study our model's sensitivity to the latent dimension $d$, the depth of residual feed-forward network $l$, the maximum sequence length $n^{\triangleright}$, as well as the dropout ratio $\rho$. For each test, based on the standard setting $\{d=64, l=1, n^{\triangleright}=20, \rho=0.6\}$, we vary the value of one hyperparameter while keeping the others unchanged, and record the new prediction result achieved. To show the performance differences, we demonstrate HR$@$10 for ranking, AUC for classification, and MAE for regression. Figure \ref{Figure:paramsensitivity} lays out the results with different parameter settings.

\textbf{Impact of $d$}.
The value of the latent dimension $d$ is examined in $\{8,16,32,64,128\}$. As an important hyperparameter in deep neural networks, the latent dimension is apparently associated with the model's expressiveness. In general, SeqFM benefits from a relatively larger $d$ for all types of tasks, but the performance improvement tends to become less significant when $d$ reaches a certain scale (32 and 64 in our case). It is worth mentioning that with $d=16$, SeqFM still outperforms nearly all the baselines in the temporal prediction tasks, which further proves the effectiveness of our proposed model.

\textbf{Impact of $l$}.
We study the impact of the depth of our shared residual feed-forward network with $l \in \{1,2,3,4,5\}$. For regression task, there is a slight performance growth for SeqFM as $l$ in creases. Though stacking more deep layers in the neural network may help the model yield better performance in some specific applications, for both ranking and classification tasks, SeqFM generally achieves higher prediction results with a smaller $l$. This is because deeper networks bring excessive parameters that can lead to overfitting, and the information learned by deeper layers may introduce noise to the model. 

\textbf{Impact of $n^{\triangleright}$}.
As can be concluded from Figure \ref{Figure:paramsensitivity}, SeqFM behaves differently on varied datasets when the maximum sequence length $n^{\triangleright}$ is adjusted in $\{10,20,30,40,50\}$. This is due to the characteristics of sequential dependencies in different datasets. For instance, in Gowalla and Foursquare, users tend to choose the next POI close to their current check-in location (i.e., the previous POI), thus forming sequential dependencies in short lengths. As a result, a larger $n^{\triangleright}$ will take more irrelevant POIs as the input, and eventually causes the performance decrease. In contrast, in Taobao, users' clicking behavior is usually motivated by their intrinsic long-term preferences, so a relatively larger $n^{\triangleright}$ can help the model achieve better results in such scenarios.

\textbf{Impact of $\rho$}.
The impact of different dropout ratios is investigated via $\rho \in \{0.5,0.6,0.7,0.8,0.9\}$. Overall, the best prediction performance of SeqFM is reached when $\rho$ is between 0.6 and 0.8. From Figure \ref{Figure:paramsensitivity} we can draw the observation that a lower dropout ratio is normally useful for preserving the model's ability to generalize to unseen test data (e.g., Foursquare and Trivago). However, on some datasets, a smaller $\rho$ comes with lower performance (e.g., Taobao and Beauty) because too many blocked neurons may result in underfitting during training.

\subsection{Importance of Key Components (RQ3)}
To better understand the performance gain from the major components proposed in SeqFM, we conduct ablation test on different degraded versions of SeqFM. Each variant removes one key component from the model, and the corresponding results on three tasks are reported. Table \ref{table:ablation} summarizes prediction outcomes in different tasks. Similar to Section~\ref{sec:paramterimpact}, HR$@$10, AUC and MAE are used. In what follows, we introduce the variants and analyze their effect respectively.

\begin{table}[t]
\vspace{-0.4cm}
\small
\caption{Ablation test with different model architectures. Numbers in bold face are the best results for corresponding metrics, and ``$\downarrow$" marks a severe (over $5\%$) performance drop.}
\vspace{-0.3cm}
\centering
\renewcommand{\arraystretch}{1.0}
\setlength\tabcolsep{1.3pt}
  \begin{tabular}{|c|c c||c c||c c|}
    \hline
     \multirow{2}{*}{Architecture} & \multicolumn{2}{c||}{HR$@$10} & \multicolumn{2}{c||}{AUC} & \multicolumn{2}{c|}{MAE}\\
    
    \cline{2-7}
 	& Gowalla & Foursquare & Trivago & Taobao & Beauty & Toys\\
    
    \hline
    Default & \textbf{0.467} & \textbf{0.431} & \textbf{0.957} & \textbf{0.826} & \textbf{0.890} & \textbf{0.704}\\
    Remove SV & 0.455 & 0.420 & { }0.892$\downarrow$ & { }0.765$\downarrow$ & { }0.959$\downarrow$ & { }0.762$\downarrow$\\
    Remove DV & { }0.424$\downarrow$ & { }0.396$\downarrow$ & { }0.862$\downarrow$ & { }0.731$\downarrow$ & { }0.972$\downarrow$ & { }0.772$\downarrow$\\
    Remove CV & { }0.430$\downarrow$ & { }0.404$\downarrow$ & 0.963 & { }0.754$\downarrow$ & { }0.935$\downarrow$ & { }0.763$\downarrow$\\
    Remove RC & 0.457 & \textbf{0.431} & { }0.898$\downarrow$ & { }0.761$\downarrow$ & 0.918 & 0.719\\
    Remove LN & 0.461 & 0.423 & 0.933 & 0.798 & 0.922 & 0.720\\
     \hline
    \end{tabular}
\label{table:ablation}
\vspace{-0.7cm}
\end{table}

\textbf{Remove Static View (Remove SV)}. The attention head in the static view models the interactions among all the static features. After removing it, a noticeable performance drop has been observed, especially on classification and regression tasks. In our application of SeqFM, the static view directly models interaction between the user and the target object (i.e., POI, link, and item), which is rather important especially when the task relies on mining users' personal preferences (e.g., the rating prediction task).

\textbf{Remove Dynamic View (Remove DV)}. The modelling of the sequential interactions among dynamic features is crucial to the model's performance in temporal predictive analytics. Hence, a significant (over 5\%) performance decrease has appeared in all three tasks. The results verify that the sequence-awareness plays a pivotal role when prediction tasks involve dynamic features. Specifically, the most severe performance drop is exerted in the classification task, including a 10\% decrease on Trivago and 12\% decrease on Taobao. As these two datasets record users' clicking behaviors on the product links provided, the entire dynamic feature sequence carries the long-term preference of each user. So, considering the dynamic dependencies can actually help our model accurately capture the rich information from the dynamic features, and eventually yield competitive prediction effectiveness. 

\textbf{Remove Cross View (Remove CV)}. Similar to the effect of discarding the dynamic view, SeqFM suffers from the obviously inferior performance (over 5\% drop) regarding 5 datasets after the cross view with self-attention head is removed. Apparently, in this degraded version of SeqFM, the interactions between static features and dynamic features are discarded, leading to a significant loss of information. This verifies the contribution of the self-attention head in the cross view to our model's final performance in all three tasks.

\textbf{Remove Residual Connections (Remove RC)}. Without residual connections, we find that the performance of SeqFM gets worse, especially on Trivago and Taobao datasets. Presumably this is because information in lower layers (i.e., the output generated by the attention head) cannot be easily propagated to the final layer, and such information is highly useful for making predictions, especially on datasets with a large amount of sparse features.

\textbf{Remove Layer Normalization (Remove LN)}. The layer normalization operation is introduced mainly for the purpose of stabilizing the training process by scaling the input with varied data scales for deep layers. Removing the layer normalization also shows a negative impact on the prediction performance, especially in the regression task where the properly normalized features can usually generate better results.  

\subsection{Training Efficiency and Scalability (RQ4)}
\begin{figure}[!t]
\vspace{-0.5cm}
\center
\includegraphics[width = 2.7in]{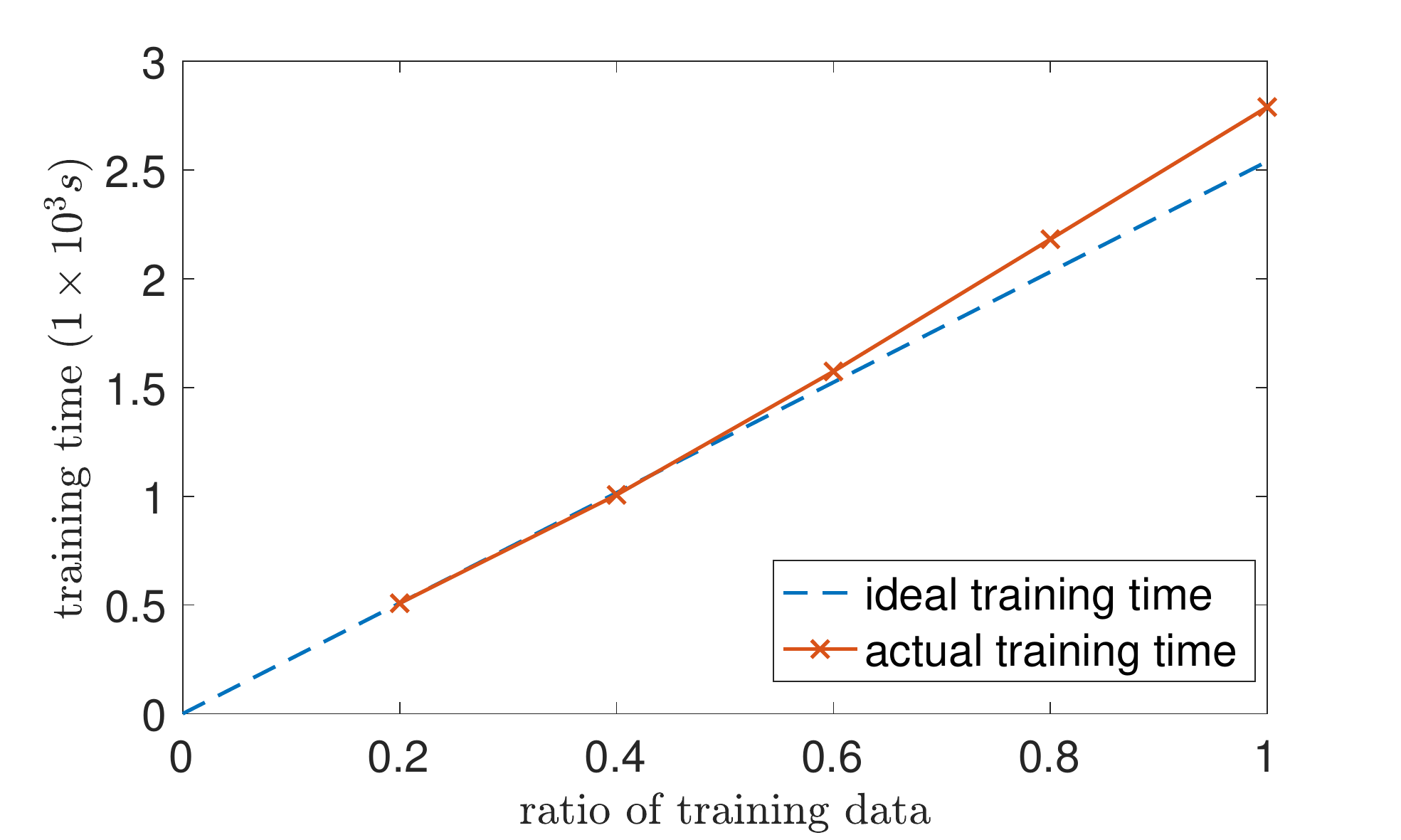}
\vspace{-0.4cm}
\caption{Training time of SeqFM w.r.t varied data proportions.}
\label{Figure:train}
\vspace{-0.7cm}
\end{figure}

We test the training efficiency and scalability of SeqFM by varying the proportions of the training data in $\{0.2, 0.4, 0.6, 0.8, 1.0\}$, and then report the corresponding time cost for the model training. It is worth noting that the Trivago dataset is used for scalability test since it contains the most instances. The growth of training time along with the data size is shown in Figure \ref{Figure:train}. When the ratio of training data gradually extends from 0.2 to 1.0, the training time for SeqFM increases from $0.51\times10^{3}$ seconds to $2.79\times10^{3}$ seconds. It shows that the dependency of training time on the data scale is approximately linear. Hence, we conclude that SeqFM is scalable to even larger datasets.

\section{Related Work}\label{sec:related}
In a nutshell, the ultimate goal of predictive analytics is to learn an effective predictor that accurately estimates the output according to the input features, where classic predictive methods like support vector machines (SVMs) \cite{chang2011libsvm} and logistic regression (LR) \cite{hosmer2013applied} have gained extensive popularity. Distinct from the continuous raw features from images and audios, features from the web-scale data are mostly discrete and categorical \cite{he2017neural}, and are therefore represented by high-dimensional but sparse one-hot encodings. When performing predictive analytics under the sparse setting, it is crucial to account for the interactions between features \cite{shan2016deep}. With the core idea of modelling high-order interactions among features, factorization machines (FMs) \cite{rendle2010factorization} are widely used for predictive analytics on very sparse data where SVMs fail. Other linear FM-based models are proposed, such as CoFM \cite{hong2013co}, field-aware FM \cite{juan2016field} and importance-aware FM \cite{oentaryo2014predicting}. However, as stated in many literatures \cite{qu2016product,lian2018xdeepfm,he2017neural,xiao2017attentional}, these models show limited effectiveness in mining high-order latent patterns or learning quality feature representations.

Another line of research on FM-based models for predictive analytics incorporates deep neural networks (DNNs)\cite{zhang2016deep,he2017neural,qu2016product,lian2018xdeepfm}. For example, the FM-supported neural network (FNN) \cite{zhang2016deep} as well as the neural factorization machine (NFM) \cite{he2017neural} are proposed to learn non-linear high-order feature interactions. They both use the pre-trained factorization machines for feature embedding before applying DNNs. Qu \textit{et al.} proposes a product-based neural network (PNN) \cite{qu2016product}, which introduces a product layer between embedding layer and DNN layer, and does not rely on pre-trained FM parameters. More recently, hybrid architectures are introduced in Wide\&Deep \cite{cheng2016wide}, DeepFM \cite{guo2017deepfm} and xDeepFM \cite{lian2018xdeepfm} by combining shallow components with deep ones to capture both low- and high-order feature interactions. Unfortunately, as discussed in Section \ref{sec:intro}, existing FM-based models lack the consideration of sequential dependencies within feature interactions, and most of them ignore the temporal orders within the data.  
On the contrary, our SeqFM extends FM-based models to the temporal predictive analytics, and utilizes the sequential dependencies within dynamic features to yield superior prediction performance.

\section{Conclusion}\label{sec:conclusion}
In this paper, we propose SeqFM, a sequence-aware factorization machine for temporal predictive analytics. For the first time, we incorporate sequential dependencies into FM-based models by proposing a novel multi-view self-attention scheme to model the interactions between different features. SeqFM is then successfully applied to three different temporal prediction tasks including ranking, classification and regression. The experimental results showcase that SeqFM is a powerful yet general model that can yield superior performance in a wide range of real-world applications.
\vspace{-0.1cm}

\section*{Acknowledgment}
This work is supported by Australian Research Council (Grant No. DP190101985, DP170103954 and DP160104075).

%\bibliographystyle{IEEEtran}
%\bibliography{IEEE}


\end{document}